\documentclass[10pt,twocolumn,letterpaper]{article}

\usepackage[pagenumbers]{cvpr} 

\usepackage{url}
\usepackage{booktabs}       
\usepackage{multirow}
\usepackage{epsfig}
\usepackage{marvosym}
\usepackage{threeparttable}
\usepackage{algorithm}
\usepackage{algorithmic}

\usepackage{bm}
\usepackage{makecell}
\usepackage{amsmath}
\usepackage{mathrsfs}
\usepackage{graphicx}
\usepackage{natbib}
\usepackage{color}
\usepackage{amsthm}
\usepackage{amsfonts}
\usepackage[most]{tcolorbox}
\usepackage{colortbl}

\newcommand{\myformer}{VITA }

\newcommand{\Rmnum}[1]{\expandafter\@slowromancap\romannumeral #1@}

\definecolor{cvprblue}{rgb}{0.21,0.49,0.74}
\usepackage[pagebackref,breaklinks,colorlinks,allcolors=cvprblue]{hyperref}

\def\paperID{1605}
\def\confName{CVPR}
\def\confYear{2026}

\title{Unifying Perception and Action: A Hybrid-Modality Pipeline with Implicit Visual Chain-of-Thought for Robotic Action Generation}

\author{Xiangkai Ma, Lekai Xing, Han Zhang, Wenzhong Li\textsuperscript{\Letter}, Sanglu Lu\\
State Key Laboratory for Novel Software Technology, Nanjing University\\
Nanjing, China\\
{\tt\small {xiangkai.ma,502024330053,zhanh}@smail.nju.edu.cn, {sanglu,lwz}@nju.edu.cn}
\thanks{
The corresponding author is Wenzhong Li.
}\\
\url{https://vita-cvpr26.github.io/}
}

\begin{document}


\twocolumn[{
\renewcommand\twocolumn[1][]{#1}
\vspace{-50pt}
\maketitle
\begin{center}
\vspace{-30pt}
\captionsetup{type=figure}
\includegraphics[width=0.9\linewidth]{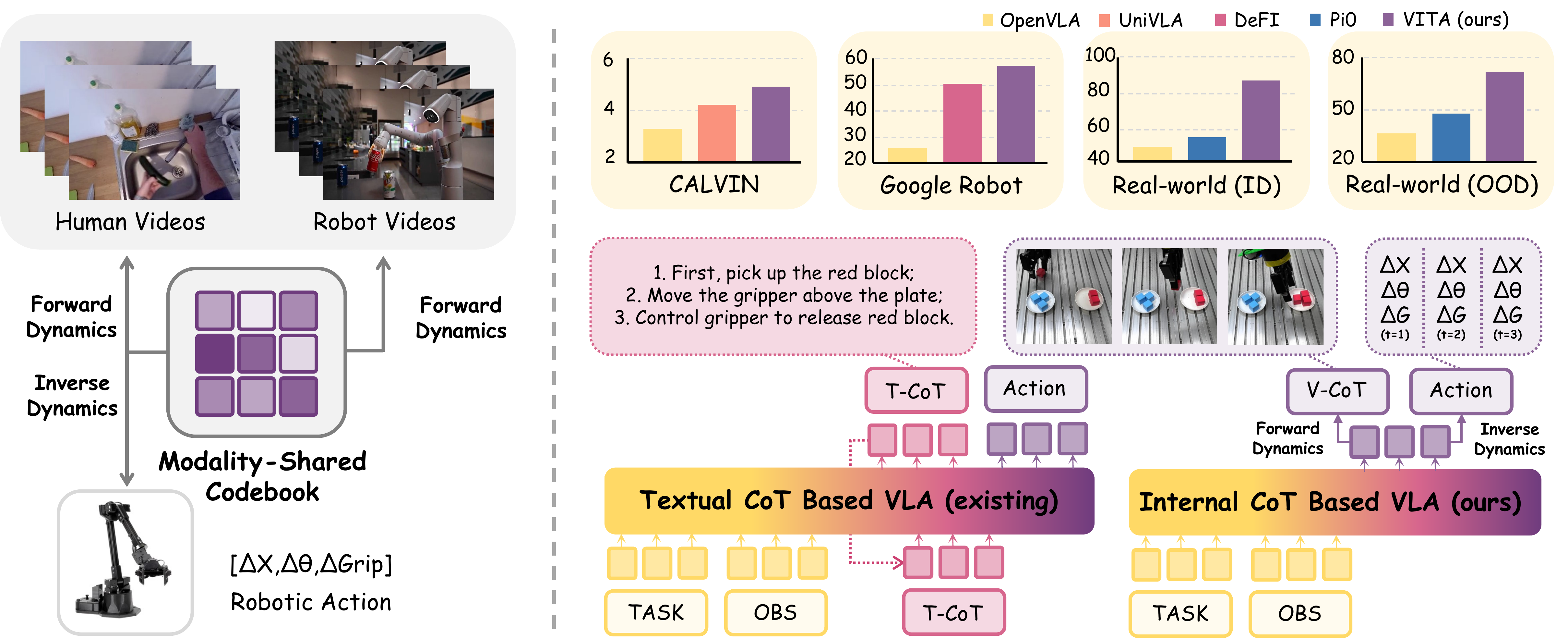}
\vspace{-1pt}
\caption{
We propose \textbf{VITA}, a novel framework that unifies visual perception and action generation. 
A cross-modal shared codebook is established, where latent variables are decoded into videos or motion trajectories through forward and inverse dynamics processes respectively. 
This dual consistency at both the representation level and optimization objectives enables VITA to effectively learn motion knowledge from extensive human demonstrations and robot operation videos.
}\label{fig:abstract}
\end{center}
}]

\begin{abstract}
Vision-Language-Action (VLA) models built upon Chain-of-Thought (CoT) have achieved remarkable success in advancing general-purpose robotic agents, owing to its significant perceptual comprehension. 
Recently, since text-only CoT struggles to adequately capture scene details in complex spatial environments, a highly promising strategy involves leveraging visual priors to guide robotic action generation.
Nevertheless, these strategies face two inherent challenges: (i) a modality gap between visual observations and low-level actions, and (ii) unstable training due to competing objectives between visual prediction and action generation. To address these challenges, we propose a Vision-Integrated Trajectory Alignment (VITA) framework that learns a shared discrete latent space for vision and action, enabling joint modeling of perception and motor control.
VITA introduces a implicit visual CoT: autoregressively generated tokens is simultaneously decoded into future frames predictions and robot actions, thereby internalizing visual dynamics as an inductive bias for motion planning. 
Extensive experiments on simulated and real-world environments demonstrate state-of-the-art performance. 
VITA improves 14.5\%, 9.6\% and 12.1\% over existing baselines on CALVIN, LIBERO and SimplerEnv. Furthermore, VITA attains an average success rate of 80.5\% across six real-world tasks, demonstrating its potential as a generalist robotic manipulation model.

\end{abstract}
\vspace{-1pt}

\section{Introduction} \label{section:introduction}
Recent research on building Vision-Language-Action (VLA) models~\cite{zitkovich2023rt2,kim2024openvla,black2024pi_0} based on Vision-Language Models (VLMs)~\cite{beyer2024paligemma,karamcheti2024prismatic,yang2025qwen3} has gained significant attention. 
These studies~\cite{qu2025spatialvla,liang2025discrete} incorporate policy head~\cite{intelligence2025pi_0.5,driess2025pi_0.5ki} or discrete action decoders~\cite{pertsch2025pi_fast} to connect visual and perception semantic prior knowledge from  VLMs, with executable motor commands. 
Recently, robotics community has focused on developing general-purpose robotic foundation models, which faces two core challenges~\cite{zheng2025x-vla}: (1) understanding human instructions in context with observed scenes, and (2) executing high-precision actions across broad real-world environments and agent embodiments.

To address these challenges, existing VLAs~\cite{luo2025learning,wu2023unleashing} involves pretraining models on extensive website multimodal datasets to achieve strong generalization. 
Early studies~\cite{zawalski2024robotic,sun2024emma} attempted to decompose high-level user instructions into a series of sub-tasks, treating them as a single-modality Chain of Thought (CoT), thereby enhancing the long-context action reasoning ability and interpretability. 
However, text-only CoT~\cite{belkhale2024rth,lin2025onetwovla} fails to fully comprehend fine-grained visual context. In the complex real-world scenes, language descriptions alone are often insufficiently grounded and semantically ambiguous. 
Consequently, a highly promising strategy leverages visual dynamics as prior to guide robot action generation, with the following avenue: 
These methods~\cite{bu2024towards,hu2024video,zhao2025cotvla,feng2025generalist,zhang2025dreamvla} utilize abundant videos to pre-train VLMs by providing initial frames and instructions, requiring model to autoregressively predict future frames. 
Subsequently, these priors are transferred to robotic manipulation through fine-tuning of the VLM and action expert. 

Nevertheless, these strategies face two inherent challenges: (i) the inherent modality gap between high-dimensional visual observations and low-level actions results in the majority of pixel-level details in generated future images being irrelevant to action execution. Consequently, directly generating precise robotic actions from observations remains difficult; (ii) The competition between optimization objectives of vision-based prediction proxy tasks and action generation tasks leads to training instability. This misalignment further prevents the action policy from fully leveraging the rich visual dynamics learned by the VLM backbone, ultimately limiting overall performance. 
Furthermore, the ``predict-then-act'' reasoning paradigm of first predicting images and then generating actions incurs significant computational latency, rendering models unsuitable for high-frequency manipulations. 
In contrast, humans do not need to mentally simulate a full trajectory of future visual states before executing an action~\cite{mazzaglia2025hybrid}. 
The brain develops skilled motor intuition based on task requirements and visual percepts, which then directly guides the cerebellum to produce precise motor commands~\cite{simon1992explanation,kahneman2009conditions}. This inspires us to unify the two processes to learning motor intuition.

To address the aforementioned challenges, we establish VITA (Vision-Integrated Trajectory Alignment) as s unified framework.
By constructing a modality-shared representation space for vision and action, VITA explicitly bridge the modality gap between perception and control. Simultaneously, by introducing joint optimization objectives, VITA internalizes future frames prediction (visual CoT) as an inductive bias for action generation. Concretely, the single sequence of tokens generated by the VLM backbone is simultaneously routed to two dedicated decoders, one for visual prediction and the other for motion reconstruction, serving two complementary sub-tasks. 
Through explicit coupling of visual perception and action generation, VITA establishes a unified learning paradigm for forward and inverse dynamics: the visual subgoal focuses on abstracting motion intuitions from predicted future scenes, while the action subgoal concentrates on inversely inferring potential action commands from the spatial evolution of motion states. 
This dual alignment at both the representation level and optimization objectives enables VITA to learn complementary knowledge from extensive cross-modal data, thereby establishing a scalable and general-purpose architecture.

\begin{figure*}
\begin{center}
\centerline{\includegraphics[width=2.0\columnwidth]{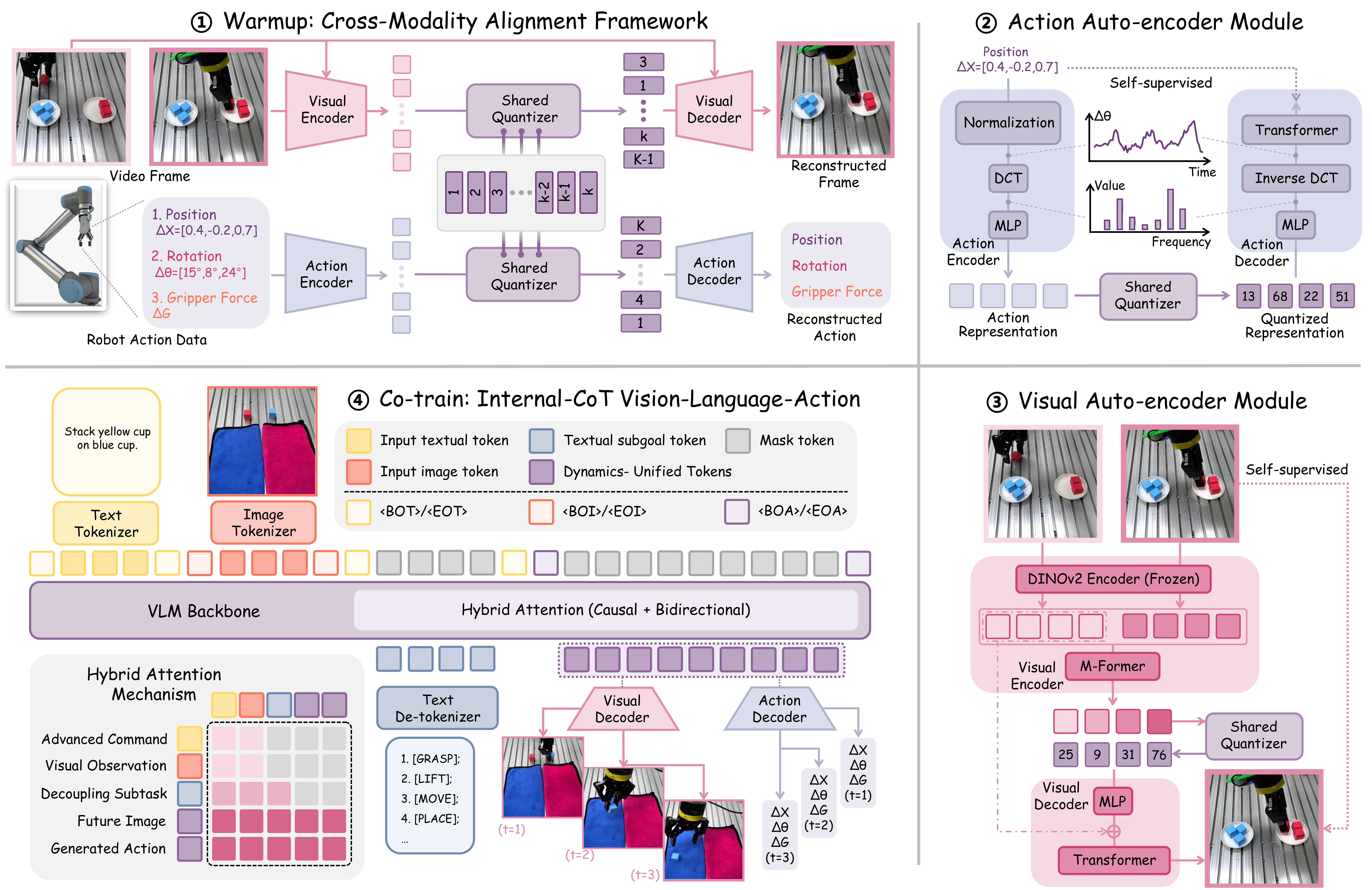}}
\vspace{-5pt}
\caption{
Overview of the \myformer framework. 
Utilizing the cross-modal alignment in \textcircled{1}, visual perception and motor control modalities are unified in the shared discrete latent space, where the dual-autoencoder architectures are illustrated in \textcircled{2} and \textcircled{3}.
Benefiting from the representation alignment, the VLM backbone in \textcircled{4} generates dynamics-unified tokens via a hybrid attention mechanism. These tokens are decoded into future frames and robot actions, as Internal CoT.
}\label{fig:method1}
\end{center}
\vspace{-25pt}
\end{figure*}

Specifically, VITA adopted a ``warmup then co-training'' regimen.
In the warmup stage, we self-supervisedly train visual and action autoencoder and a modal-shared quantizer, where encoded vectors in codebook form the unified discrete latent space. 
For the visual modality, we extract and quantize visual motion-aware features from consecutive frames and use them to reconstruct future frames as the optimization objective. 
For the action modality, we adopt a frequency-aware temporal encoding strategy inspired by FAST~\cite{pertsch2025pi_fast}. 
Crucially, the visual and action branches are trained independently, requiring no cross-modal aligned pairs.
In the co-trainstage, each token generated by the VLM backbone can be mapped to an encoding vector in the common representation space. 
This vector is then processed through two independent decoding pipelines to yield predictions of future images and corresponding robot action sequences.
In the evaluation stage, we only retained the VLM backbone and action decoder as the lightweight architecture, to reduce the inference latency.

Our contributions can be summarized as follows. 
\begin{itemize}
    \item We propose VITA, a novel VLA framework that aligns perception and action via a unified latent space, and internalizes future frames prediction as an inductive bias for action generation to unify forward and inverse dynamics.
    \item We introduce a progressive training regimen, that endows VITA with generalizable motion dynamics learned from diverse real-world interaction video, while discarding motion-irrelevant pixel-level details.
    \item VITA achieves 14.5\%, 9.6\% and 12.1\% improvements on CALVIN, LIBERO and SimplerEnv simulations, and attains an average success rate of 80.5\% across six real-world tasks, demonstrating superior generalization
\end{itemize}

\section{Related Work} \label{section:relatedwork}

\subsection{VLA Models}
By incorporating real-world prior knowledge about physical laws and social conventions, VLM-based VLAs~\cite{zitkovich2023rt2,kim2024openvla,wang2024scaling} acquire strong multimodal reasoning capabilities to generalize to entirely new observational scenarios. 
Subsequent works~\cite{team2024octo,bi2025h} have thoroughly examined strategies for action generation, including (i) autoregressive prediction of the next token~\cite{belkhale2024rth,kim2024openvla,kim2025openvla-oft,pertsch2025pi_fast} and (ii) expert policy modeling via Diffusion or Flow Matching models~\cite{team2024octo,li2023roboflamingo,li2024cogact,black2024pi_0}. Recent efforts~\cite{shi2025pi_hi} further aim to develop general-purpose VLA agents that generalize across embodiments, scenarios, and tasks. To this end, existing methods\cite{zawalski2024robotic,chen2025training,duan2025fast} have introduced chain-of-thought to enhance VLAs' contextual understanding and long-horizon task reasoning capabilities, achieving significant success.

Nevertheless, these text-only CoT approaches lack the ability to perceive spatial topological features from visual observations and anticipate scene dynamics~\cite{huang2025tactile}. To address the challenge, VITA enhances VLAs' spatial cognition and dynamic prediction in complex environments, enabling more grounded and robust action generation.

\subsection{Visual-guided VLAs}
Video models~\cite{liu2024sora,agarwal2025cosmos,wan2025wan} have demonstrated excellent performance in producing physically plausible and temporally coherent videos. Some researchers have attempted to leverage the physical principles, motion patterns, and spatiotemporal priors to develop general-purpose VLA architectures. Early approaches, such as SuSIE~\cite{black2023susie} and UniPi~\cite{du2023unipi}, modeled inverse dynamics by extracting action signals from predicted future frames. GR series~\cite{wu2023gr1,cheang2024gr2,li2025gr3} pre-trained on human demonstration videos to learn transferable robotic policies. Methods like Moto~\cite{chen2024moto}, Seer~\cite{tian2024seer}, and LAPA~\cite{ye2024lapa} enhanced VLA reasoning by predicting future images to form visual chains-of-thought. Concurrently, CoT-VLA~\cite{zhao2025cotvla} established an explicit visual CoT, following the paradigm of ``predict-then-act'', to improve action generation. UniVLA~\cite{wang2025univla} and DeFI~\cite{pretraining2025defi} utilize large-scale unlabeled human demonstration videos to learn general-purpose action representations.

However, existing methods isolate visual and action generation into two separate streams, where differences in optimization objectives can lead to rapid forgetting of pre-trained knowledge. Furthermore, the inherent modality gap makes directly aligning visual inputs with output commands a challenging task. 
To overcome these issues, VITA aims to integrate visual prediction and action reasoning into a consistent process, enhancing the reasoning capability by capturing representation of forward dynamics. Furthermore, VITA explicitly bridges the vision–action modality gap, enabling seamless knowledge transfer from video-based pretraining to precise motor execution.

\vspace{-5pt}
\section{\myformer Architecture} \label{section:vita}

We introduce VITA (Vision-Integrated Trajectory Alignment), a unified framework for VLA modeling that bridges the semantic gap between high-dimensional visual observations and low-level robotic actions through cross-modal latent representation learning. 
Furthermore, we implicitly establishes ``visual chain-of-thought'' reasoning: the model learns to anticipate scene evolution while deriving action policies. 
This is similar to the human's motor intuition where perceptual anticipation directly informs motor execution without explicit trajectory simulation.
Specifically, our training proceeds in three stages:

\noindent
(1) \textit{Warmup}: We mix video data and action trajectories at the batch level. Each batch contains single-modality samples, and we self-supervised train the modality-specific modules and the shared codebook via reconstruction losses.

\noindent
(2) \textit{Co-train}: After freezing the trained codebook, we attach the visual and action decoders to the output of the VLM backbone. During the co-train stage, we continue training the VLM backbone along with both decoders. We use synchronized robot operation videos and action trajectories as supervision to compute the loss against the outputs of the two decoders, as shown in Figure 1.

\noindent
(3) \textit{Fine-tune}: Finally, we fine-tune only the action decoder on specific simulated and real-world datasets, while keeping the VLM backbone frozen.

\subsection{Cross-modal Vector Quantization Framework}
To bridge the inherent modality mismatch between pixel-rich visual inputs and sparse action outputs, we construct a cross-modal vector quantization framework that projects both modalities into a unified discrete representation space. 
Given a shared codebook \begin{small}$\mathcal{C}=\{c_k\}_{k=1}^K\subset\mathbb{R}^d$\end{small}, the quantization process is defined as follows. 

Let \begin{small}$\mathcal{Q}{:}\mathbb{R}^d\to\mathcal{C}$\end{small} denotes the vector quantization operator:
\begin{small}
\begin{equation}
\vspace{-5pt}
  \begin{split}
    \mathcal{Q}(z)=c_k,~where~k=\arg\min_j\left\|z-c_j\right\|_2.
  \end{split}
\vspace{-15pt}
\end{equation}
\end{small}

During warmup, visual and action encoding modules \begin{small}$(\mathcal{E}_v,\mathcal{E}_a)$\end{small} are trained independently to produce quantizable embeddings that minimize reconstruction loss under their respective decoders \begin{small}$(\mathcal{D}_v,\mathcal{D}_a)$\end{small}, respectively. 
Therefore, cross-modal paired supervision is not required at this stage. 
Moreover, since both modalities share the same quantizer and codebook, partial gradients during the reconstruction process jointly optimize the quantization components.


\subsubsection{Visual Auto-encoder and Quantization Module}
The visual module processes consecutive video frames to extract motion-aware representations. Given a pair of consecutive frames \begin{small}$(\mathbf{I}_t,\mathbf{I}_{t+1})$\end{small}, the visual encoder \begin{small}$\mathcal{E}_{v}$\end{small} first extracts spatial-temporal dense features via DINOv2-encoder:
\begin{small}
\begin{equation}
  \begin{split}
    f_t=DINOv2(\mathbf{I}_t),~f_{t+1}=DINOv2(\mathbf{I}_{t+1}), \\
    z_v=M\textit{-}Former([f_t;f_{t+1}])\in\mathbb{R}^d,
  \end{split}
\end{equation}
\end{small}
Where \begin{small}$\left[;\right]$\end{small} denotes channel-wise concatenation and the memory-augmented M-Former is utilized to yield a compact spatial-temporal motion embedding. The quantized visual latent is then obtained as:
\begin{small}
\begin{equation}
\vspace{-5pt}
  \begin{split}
    \hat{z}_v=\mathcal{Q}(z_v),
  \end{split}
\vspace{-5pt}
\end{equation}
\end{small}
and the quantized latent vectors are utilized to reconstruct the future frame via the visual decoder:
\begin{small}
\begin{equation}
  \begin{split}
    \hat{\mathbf{I}}_{t+1}=\mathcal{D}_{v}(f_{t},\hat{z}_{v}).
  \end{split}
\end{equation}
\end{small}
The visual reconstruction loss is calculated by L1 loss~\cite{hastie2009l1_loss} and SSIM loss~\cite{wang2004ssim_loss}:
\begin{small}
\begin{equation}\label{eq:lv}
  \begin{split}
    \mathcal{L}_v=\lambda_{L1}\left\|\mathbf{I}_{t+1}\text{-}\hat{\mathbf{I}}_{t+1}\right\|_1+\lambda_{SSIM}(1\text{-}SSIM(\mathbf{I}_{t+1},\hat{\mathbf{I}}_{t+1})).
  \end{split}
\end{equation}
\end{small}

Crucially, the quantizer is trained only on visual data, without any action supervision, allowing it to learn rich motion priors from the vast amount of internet human demonstration and robot operation videos.

\subsubsection{Action Auto-encoder and Quantization Module}
For action sequences, we normalize each control dimension (position \begin{small}$\Delta x$\end{small}, rotation \begin{small}$\Delta \theta$\end{small}, gripper force \begin{small}$\Delta F$\end{small}). 
Specifically, we denote \begin{small}$\mathbf{a}_{t:t+H}=[\mathbf{a}_t,\mathbf{a}_{t+1},...,\mathbf{a}_{t+H-1}]^{\top}\in\mathbb{R}^{H\times d_a}$\end{small} as an action segment of horizon \begin{small}$H$\end{small}, where each \begin{small}$\mathbf{a}_\tau\text{=}[\Delta x_\tau,\Delta\theta_\tau,\Delta F_\tau]^{\top}$\end{small}. 
Then these action data are compressed temporal dynamics into frequency-domain coefficients using Discrete Cosine Transform (DCT)~\cite{ahmed2006dct}, followed by a lightweight MLP to encode coefficients into action representations as:
\begin{small}
\begin{equation}
\vspace{-5pt}
  \begin{split}
    \hat{\mathbf{a}}=DCT(\mathbf{a}_{t:t+H})\in\mathbb{R}^{H\times d_a}, \\
    z_a=MLP(flatten(\hat{\mathbf{a}}))\in\mathbb{R}^d.
  \end{split}
  \vspace{-10pt}
\end{equation}
\end{small}

These features are then quantized using the shared codebook \begin{small}$\mathcal{C}$\end{small}, ensuring semantic alignment with visual tokens:
\begin{small}
\begin{equation}
\vspace{-5pt}
  \begin{split}
    \hat{z}_a=\mathcal{Q}(z_a).
  \end{split}
  \vspace{-15pt}
\end{equation}
\end{small}

The quantized action latent vectors are decoded back into continuous reconstructed action trajectory via:
\begin{small}
\begin{equation}
\vspace{-5pt}
  \begin{split}
    \hat{\mathbf{a}}_{t:t+H}=MLP(InverseDCT(\hat{z}_a)),
  \end{split}
\end{equation}
\end{small}
and this process is defined as \begin{small}$\hat{\mathbf{a}}_{t:t+H}=\mathcal{D}_a(\hat{z}_a)$\end{small}.
The action encoding and quantization are supervised by MSE loss against ground-truth actions:
\begin{small}
\begin{equation}\label{eq:la}
\vspace{-5pt}
  \begin{split}
    \mathcal{L}_a=\left\|\mathbf{a}_{t:t+H}-\hat{\mathbf{a}}_{t:t+H}\right\|_2^2.
  \end{split}
  \vspace{-15pt}
\end{equation}
\end{small}

Our key design insight lies in sharing the codebook between visual and action modalities, enabling VITA to achieve structural consistency in the latent space. This implicit alignment facilitates downstream joint optimization without requiring explicit cross-modal annotations.

\subsection{VLM Backbone Architecture}
The VLM backbone implements a two-stage reasoning process that decouples high-level instruction following from low-level action generation, thereby enabling interpretable and temporally coherent policy execution. This process consists of textual and internal CoT Reasoning.

\subsubsection{Progressive Attention Mechanism}
To coordinate these two reasoning streams, we design a progressive attention mechanism, as illustrated in Figure~\ref{fig:method1}~\textcircled{4}. Specifically, we refer to the tokens derived from the input text instruction and the current visual observation as input tokens. The tokens generated by the two chain-of-thought processes are then explicitly partitioned into two categories: textual tokens and cross-modal tokens.

During inference, we first apply bidirectional attention within the input tokens to fully capture global context. We then generate textual tokens in parallel. When generating cross-modal tokens, we apply bidirectional attention separately within the input tokens and within the textual tokens to enable comprehensive intra-chain interaction. Furthermore, we impose a causal attention between these token groups to establish a directed information flow:
\begin{small}
\begin{equation}
\vspace{-5pt}
  \begin{split}
    \mathrm{input}\to\mathrm{textual}\to\mathrm{cross\textit{-}modal}.
  \end{split}
  \vspace{-15pt}
\end{equation}
\end{small}

This progressive attention design enables VITA to structure action prediction as two collaborative yet decoupled reasoning processes:

\noindent
(1) \textit{Perception understanding via Textual CoT}: The model abstracts structured task semantics from the language instruction and visual observation, producing an interpretable subtasks that maps high-level intent to symbolic actions;

\noindent
(2) \textit{Motion planning via Internal CoT}: Guided by the subtask prior, the model coherently generates low-dimensional action commands that align with the anticipated evolution of the future visual scene, thereby bridging symbolic planning to physical execution.

\subsubsection{Textual Chain-of-Thought Reasoning}
Firstly, the backbone decomposes the high-level instruction into a sequence of symbolic subtasks.

Formally, we let \begin{small}$\mathbb{Z}_{sub}\text{=\{[GRASP],[LIFT],[MOVE],[PLACE],}$\end{small}
\begin{small}$\text{[ROTATE],[OPEN],[CLOSE],[END SUBTASK]}\}$\end{small} 
denote the fixed subtask vocabulary, where each token corresponds to a semantically meaningful action chunk commonly observed in robotic manipulation tasks.

Given the language instruction \begin{small}$\mathbf{x}$\end{small}, initial observation \begin{small}$\mathbf{I}_0\in\mathbb{R}^{H\times W\times3}$\end{small} and robot state \begin{small}$\mathbf{s}$\end{small}, we have multimodal context:
\begin{small}
\begin{equation}
\vspace{-5pt}
  \begin{split}
    h_{ctx}=[T_{text}(\mathbf{x});T_{image}(\mathbf{I}_0);\mathbf{s}]\in\mathbb{R}^{N\times d}.
  \end{split}
  \vspace{-15pt}
\end{equation}
\end{small}
The VLM backbone generates a subtask sequence \begin{small}$Z_{sub}=[z_1,z_2,...,z_M]$\end{small} with \begin{small}$z_m\in\mathbb{Z}_{sub}$\end{small} denotes a symbolic action primitive. Formally, this stage is implemented as:
\begin{small}
\begin{equation}
\vspace{-5pt}
  \begin{split}
    Z_{sub}=\arg\max_{z\in\mathbb{Z}_{sub}}p_\theta(z\mid h_{ctx}),
  \end{split}
  \vspace{-15pt}
\end{equation}
\end{small}
where \begin{small}$T_{text}(\cdot)$\end{small} and \begin{small}$T_{image}(\cdot)$\end{small} represent the textual and visual tokenizer, and \begin{small}$p_{\theta}$\end{small} denotes the parameters of the VLM. 
The symbolic subtasks are then converted into additional text token embeddings via the textual tokenizer:
\begin{small}
\begin{equation}
\vspace{-5pt}
  \begin{split}
    e_{sub}=T_{text}(Z_{sub})\in\mathbb{R}^{M\times d}.
  \end{split}
  \vspace{-15pt}
\end{equation}
\end{small}
Notably, VITA indirectly trains the generation of textual chain-of-thought through multimodal joint optimization during training, but does not employ explicit subtask label supervision. This design balances scalability, stability, and reasoning interpretability.

\subsubsection{Visual Chain-of-Thought Generation}
In the second stage, the backbone takes as input the concatenation of all context tokens and the subtask embeddings:
\begin{small}
\begin{equation}
  \begin{split}
    h_{mul}=[T_{text}(\mathbf{x});T_{image}(\mathbf{I}_0);\mathbf{s};e_{sub}]\in\mathbb{R}^{(N+M)\times d}.
  \end{split}
\end{equation}
\end{small}
It then autoregressively generates a sequence of visual-action hybrid modality latent tokens:
\begin{small}
\begin{equation}
  \begin{split}
    \{\tau_i\}_{i=1}^L\sim p_\theta(\cdot\mid h_{mul}),
  \end{split}
\end{equation}
\end{small}
where each \begin{small}$\tau_i\in\{1,2,...,K\}$\end{small} indexes the shared codebook \begin{small}$\mathcal{C}$\end{small}, and \begin{small}$L$\end{small} is the token sequence length output by VLM.

Finally, the generated token sequence \begin{small}$\{\tau_{i}\}_{i=1}^{L}$\end{small} is routed to two parallel decoders:
\begin{small}
\begin{equation}
  \begin{split}
    \hat{\mathbf{I}}_{1:T}=\mathcal{D}_{v}(\{c_{\tau_{i}}\}_{i=1}^{L}),~\hat{\mathbf{a}}_{1:H}=\mathcal{D}_{a}(\{c_{\tau_{i}}\}_{i=1}^{L}),
  \end{split}
\end{equation}
\end{small}
realizing the unified prediction of future scenes and robot actions from a single latent stream. Here, \begin{small}$T$\end{small} and \begin{small}$H$\end{small} respectively represent the visual and action prediction horizons.

\subsection{Progressive Training Regimen}
Our training protocol is designed to progressively align vision, language, and action through a curriculum that leverages diverse data sources at each stage.

\subsubsection{Warmup Stage: Modality-Shared Representations}
We train the visual module \begin{small}$(\mathcal{E}_v,\mathcal{D}_v)$\end{small}, action module \begin{small}$(\mathcal{E}_a,\mathcal{D}_a)$\end{small}, and codebook \begin{small}$\mathcal{C}$\end{small} on independent video and action data.

Crucially, no cross-modal alignment (e.g., synchronized video-action pairs) is required in the warmup stage. The visual branch is trained via frame prediction:
\begin{small}
\begin{equation}
  \begin{split}
    \min_{\mathcal{E}_v,\mathcal{D}_v,\mathcal{C}}\mathbb{E}_{\mathbf{I}_t,\mathbf{I}_{t+1}}[\mathcal{L}_v],
  \end{split}
\end{equation}
\end{small}
while the action branch learns to reconstruct action sequences from quantized latents:
\begin{small}
\begin{equation}
  \begin{split}
    \min_{\mathcal{E}_{a},\mathcal{D}_{a},\mathcal{C}}\mathbb{E}_{a_{t:t+H}}[\mathcal{L}_a].
  \end{split}
\end{equation}
\end{small}
Here, \begin{small}$\mathcal{L}_v$\end{small} and \begin{small}$\mathcal{L}_a$\end{small} are defined in formula~\ref{eq:lv} and~\ref{eq:la}.
This stage establishes a modality-agnostic latent vocabulary that enables downstream cross-modal token sharing.

\subsubsection{Co-train Stage: Aligning Visual Priors with Action}
In the co-train stage, we jointly leverage two types of data:
(1) Only video data (e.g., human demonstration videos without action annotations);
(2) Synchronized vision-action paired data (i.e., robot operation videos with aligned robot motion trajectories). During this stage, we train the VLM backbone, visual decoder \begin{small}$\mathcal{D}_v$\end{small}, and action decoder \begin{small}$\mathcal{D}_a$\end{small}, using a mixed data stream. And we freeze \begin{small}$\mathcal{C}$\end{small} from warmup. Each training batch may contain either type of data:

\noindent
(1) If a sample includes only video observations \begin{small}$(\mathbf{I}_{0:T},\mathbf{x})$\end{small} without ground-truth actions, Our VITA predicts a sequence of visual tokens \begin{small}$\{\tau_i\}_{i=1}^L$\end{small}, where each token \begin{small}$\tau_i\in\{1,2,...,K\}$\end{small} indexes a shared codebook \begin{small}$\mathcal{C}=\{c_{k}\}_{k=1}^{K}$\end{small}. These tokens are then decoded into a sequence of future images \begin{small}$\hat{\mathbf{I}}_{1:T}=\mathcal{D}_{v}(\{c_{\tau_{i}}\}_{i=1}^{L})$\end{small}. The loss is purely visual:
\begin{small}
\begin{equation}
  \begin{split}
    \mathcal{L}_{\mathrm{co}}=\sum_{t=1}^T\left(\lambda_{\mathrm{L}1}\left\|\mathbf{I}_t\text{-}\hat{\mathbf{I}}_t\right\|_1+\lambda_{\mathrm{SSIM}}\left(1\text{-}\mathrm{SSIM}(\mathbf{I}_t,\hat{\mathbf{I}}_t)\right)\right),
  \end{split}
\end{equation}
\end{small}

\noindent
(2) If a sample includes synchronized vision-action pairs \begin{small}$(\mathbf{I}_{0:T},\mathbf{x},\mathbf{a}_{1:H})$\end{small}, the model now generates a unified token sequence \begin{small}$\{\tau_i\}_{i=1}^L$\end{small}. 
These tokens are then simultaneously decoded into two parallel outputs: future images \begin{small}$\hat{\mathbf{I}}_{1:T}=\mathcal{D}_v(\{c_{\tau_i}\}_{i=1}^L)$\end{small}, and action trajectories \begin{small}$\hat{\mathbf{a}}_{1:H}=\mathcal{D}_a(\{c_{\tau_i}\}_{i=1}^L)$\end{small}. 
This dual-decoding paradigm enables end-to-end alignment between perception and actuation without explicit intermediate trajectory simulation. Both visual and motion reconstructions are utilized to calculate the loss function:
\begin{small}
\begin{equation}
  \begin{split}
    \mathcal{L}_{\mathrm{co}}=\lambda_v\underbrace{\|\mathbf{I}_{1:T}-\hat{\mathbf{I}}_{1:T}\|_1}_{\mathrm{visual~CoT}}+\lambda_a\underbrace{\|\mathbf{a}_{1:H}-\hat{\mathbf{a}}_{1:H}\|_2^2}_{\text{action generation}}.
  \end{split}
\end{equation}
\end{small}

This mixed-objective formulation enables seamless integration of large-scale unlabeled video data with high-quality paired trajectories. Crucially, because both vision and action decoders share the same discrete latent token stream generated by the VLM backbone (via the shared quantizer), the visual prediction task provides an inductive bias that regularizes action generation, while the action supervision distills only task-relevant visual dynamics, effectively filtering out motion-irrelevant pixel details.



\section{Experiments} \label{section:experiments}
\subsection{Implementation Details}
\subsubsection{Hybrid Datasets}
We organize 13 datasets collected from diverse sources into a hybrid dataset for the warmup and co-train stages. Our collection protocol focuses on the following: 
(1) Human demonstration video datasets (SSv2~\cite{goyal2017ssv2} and Ego4D~\cite{grauman2022ego4d}); 
(2) Robot videos and action data in real-world (OXE~\cite{o2024oxe} and RoboMIND~\cite{wu2024robomind}); 
(3) Robot videos and action data in simulation (CALVIN-ABC~\cite{mees2022calvin} and LIBERO~\cite{liu2023libero}).
Details in Table 10 and Section 7.1 of appendix.

\subsubsection{Model Architecture and Training Details}
Our VITA implementation follows mainstream design of Pi0~\cite{black2024pi_0}, consisting of SigLIP~\cite{zhai2023siglip} as visual tokenizer (400M) and Gemma~\cite{team2024gemma} as backbone (2B). 
Besides, we utilize a 12-layer ViT~\cite{dosovitskiy2020vit} as visual decoder (96M) and a transformer architecture as the action decoder (228M parameters). 
The action decoder bears the core responsibility of precisely reconstructing high-dimensional and temporally coherent robot action trajectories, thus it is designed with a larger parameter capacity. 
The model is trained on 16 NVIDIA A100 GPUs. We perform 300K steps of training with 2.8B trainable parameters, requiring approximately 5 days.
Details in Table 11 and Section 8 of appendix.


\subsubsection{Benchmarks}
We evaluate VITA's performance on three simulated benchmarks: CALVIN~\cite{mees2022calvin}, LIBERO~\cite{liu2023libero}, and SimplerEnv~\cite{li2024simplerenv}. 
Details of simulation benchmark in Section 9 of appendix.
These benchmarks encompass manipulation tasks in multi-step household scenarios with varying object configurations. In the real world, we assess the model’s generalization on tabletop tasks under both in-distribution (ID) and out-of-distribution (OOD) settings. 
To ensure fair comparison, we fine-tune both VITA and baselines, on simulation dataset and real-world datasets (collected on the UR-5e robotic platform). 
Details of UR-5e dataset the Section 10.2 of appendix.

\subsubsection{Baselines}
To thoroughly evaluate the advantages of the proposed VITA, we select the most competitive generalist manipulation policies as baselines. 
Specifically, for the simulated benchmarks, we compare VITA against reported performances of GR-1~\cite{wu2023gr1}, OpenVLA~\cite{kim2024openvla}, Octo~\cite{team2024octo}, CogACT~\cite{li2024cogact}, UP-VLA~\cite{zhang2025upvla}, CoT-VLA~\cite{zhao2025cotvla}, TraceVLA~\cite{zheng2024tracevla}, SpatialVLA~\cite{qu2025spatialvla}, UniVLA~\cite{wang2025univla} and DeFI~\cite{pretraining2025defi}. 
For real-world evaluation, we adopt Pi0~\cite{black2024pi_0}, OpenVLA~\cite{kim2024openvla} and GR00T N1.5~\cite{bjorck2025gr00tn1.5} as baseline methods. 
All baseline results are directly taken from original publications or reproduced using their publicly available implementations.

\begin{table}[ht]
\vspace{-5pt}
\caption{
Results on CALVIN ABC-D. We report The average number of tasks completed after executing 5 consecutive instructions over 1,000 evaluation rollouts. 
}\label{tab:result_calvin}
\vspace{-8pt}
\centering
\resizebox{1.0\columnwidth}{!}{
\begin{small}
\renewcommand{\multirowsetup}{\centering}
\tabcolsep=0.3cm
\renewcommand\arraystretch{1.0}
\begin{tabular}{c|ccccc|c}
\toprule
\hline
\multicolumn{1}{c|}{\multirow{2}{*}{{Models}}} & 
\multicolumn{5}{c|}{Task Completed rate in a Row} & 
\multicolumn{1}{c}{\multirow{2}{*}{{Average Number}}} \\

& 1 & 2 & 3 & 4 & 5 & \\
\hline
GR-1~\cite{wu2023gr1} &             85.4 & 71.2 & 59.6 & 49.7 & 40.1 & 3.06 \\
OpenVLA~\cite{kim2024openvla} &     91.3 & 77.8 & 62.0 & 52.1 & 43.5 & 3.27 \\
Pi0~\cite{black2024pi_0} &          93.8 & 85.0 & 76.7 & 68.1 & 59.9 & 3.92 \\
UP-VLA~\cite{zhang2025upvla} &      92.8 & 86.5 & 81.5 & 76.9 & 69.9 & 4.08 \\
UniVLA~\cite{wang2025univla} &      98.9 & 94.8 & 89.0 & 82.8 & 75.1 & 4.41 \\
DeFI~\cite{pretraining2025defi} &   97.9 & 94.2 & 90.7 & 87.0 & 81.2 & 4.51 \\

\rowcolor{blue!10}
\textbf{VITA (ours)} & \textbf{99.1} & \textbf{94.9} & \textbf{91.2} & \textbf{87.8} & \textbf{84.5} & \textbf{4.73} \\

\hline
\bottomrule
\end{tabular}
\end{small}
}
\vspace{-15pt}
\end{table}
\subsection{Evaluation on the CALVIN}
As shown in Table~\ref{tab:result_calvin}, VITA achieves comprehensive state-of-the-art performance on the CALVIN ABC-D benchmark.
VITA not only significantly surpasses direct image-to-action mapping approaches such as GR-1, OpenVLA, and Pi0, but also exceeds pre-trained models leveraging human videos like UniVLA, as well as those incorporating dynamic modeling such as DeFI and UP-VLA, particularly excelling in long-horizon tasks (complete 3 to 5 tasks consecutively). 
On the Avg. Len metric, VITA achieves a score of 4.67, representing improvements of 5.9\% and 14.5\% over baselines UniVLA and UP-VLA. 
This demonstrates superior action reasoning and sequential planning capabilities in complex, long-horizon manipulation tasks. Which is attributed to VITA significantly enhancing long-range reasoning by internalizing future video frame prediction as an inductive bias within the action generation process.

\begin{figure*}
\begin{center}
\centerline{\includegraphics[width=2.0\columnwidth]{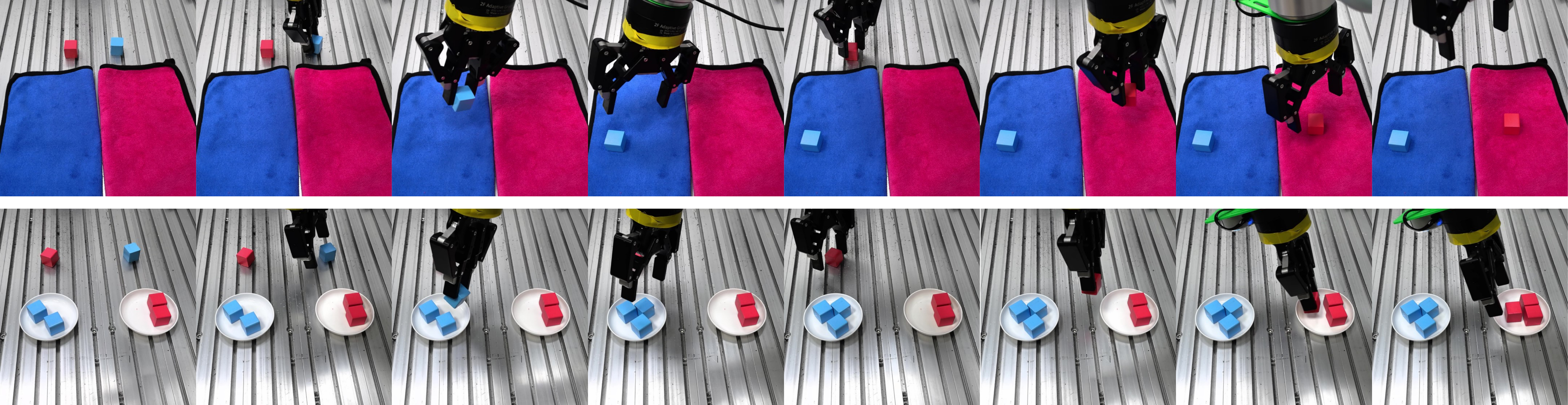}}
\caption{
Visualization of the ``contextual reasoning and color matching'' in the real world.
}\label{fig:exper3}
\end{center}
\vspace{-30pt}
\end{figure*}

\subsection{Evaluation on the LIBERO}
\begin{table}[ht]
\caption{
Performance comparison of VITA and baseline models on the LIBERO simulations.
}\label{tab:result_libero}
\vspace{-8pt}
\centering
\resizebox{1.0\columnwidth}{!}{
\begin{small}
\renewcommand{\multirowsetup}{\centering}
\tabcolsep=0.3cm
\renewcommand\arraystretch{1.0}
\begin{tabular}{c|ccccc}
\toprule
\hline
Models & GOAL & SPATIAL & OBJECT & LONG & Average \\

\hline
Octo~\cite{team2024octo} &             84.6 & 78.9 & 85.7 & 51.1 & 75.1 \\
OpenVLA~\cite{kim2024openvla} &        79.2 & 84.9 & 88.4 & 53.7 & 76.5 \\
SpatialVLA~\cite{qu2025spatialvla} &   78.6 & 88.2 & 89.9 & 55.5 & 78.1 \\
CoT-VLA~\cite{zhao2025cotvla} &        87.6 & 87.5 & 91.6 & 69.0 & 81.1 \\
Pi0-FAST~\cite{pertsch2025pi_fast} &   88.6 & 96.4 & 96.8 & 60.2 & 85.5 \\
UniVLA~\cite{wang2025univla} &         93.6 & 95.4 & 98.8 & 94.0 & 95.5 \\

\rowcolor{blue!10}
\textbf{VITA (ours)} & \textbf{95.1} & \textbf{95.9} & \textbf{98.9} & \textbf{96.8} & \textbf{96.7} \\

\hline
\bottomrule
\end{tabular}
\end{small}
}
\vspace{-15pt}
\end{table}
In Table~\ref{tab:result_libero}, VITA surpasses the strongest baseline, UniVLA, across all four LIBERO suites, demonstrating its superior capability under multitask generalization and vision-action modeling. 
On the most challenging LIBERO-Long, VITA achieves a substantial 36.2\% improvement over CoT-VLA, highlighting VITA's enhanced ability to model long-horizon task sequences and effectively overcome the performance bottlenecks of existing approaches.

\subsection{Evaluation on the SimplerEnv}
On the Google Robot branch in Table~\ref{tab:result_simplerenv_google}, VITA achieves average success rates of 57.4\%, surpassing DeFI's 51.2\%, demonstrating the efficacy of its joint vision-action co-learning. 
On the WidowX branch in Table~\ref{tab:result_simplerenv_widowx}, VITA attains an average success rate of 71.5\%, outperforming the previous best method UniVLA. 
More significantly, VITA's breakthrough is most pronounced on extreme tasks: for instance, performance on ``Pick Carrot on Plate'' jumps from OpenVLA's 20.8\% to 68.8\%. 
These results demonstrate that VITA not only achieves incremental improvements in overall performance but also establishes a robust foundation for sim-to-real transfer.
\begin{table}[ht]
\vspace{-5pt}
\caption{
Performance evaluation of VITA and baselines on the SimplerEnv-GoogleRobot benchmark (visual matching).
}\label{tab:result_simplerenv_google}
\vspace{-8pt}
\centering
\resizebox{1.0\columnwidth}{!}{
\begin{small}
\renewcommand{\multirowsetup}{\centering}
\tabcolsep=0.3cm
\renewcommand\arraystretch{1.0}
\begin{tabular}{c|cccc}
\toprule
\hline
Models & PickCokeCan & MoveNear & OpenDrawer & Average \\

\hline
Octo~\cite{team2024octo} &	        17.0 & 4.2 & 22.7 & 16.8 \\
TraceVLA~\cite{zheng2024tracevla} & 28.0 & 53.7 & 57.0 & 42.0 \\
OpenVLA~\cite{kim2024openvla} &    	16.3 & 46.2 & 35.6 & 27.7 \\
DeFI~\cite{pretraining2025defi} &	54.2 & \textbf{60.7} & 38.6 & 51.2 \\			

\rowcolor{blue!10}
\textbf{VITA (ours)} & \textbf{57.5} & 55.8 & \textbf{58.9} & \textbf{57.4} \\

\hline
\bottomrule
\end{tabular}
\end{small}
}
\vspace{-20pt}
\end{table}
\begin{table}[ht]
\caption{
Performance evaluation of VITA and existing baseline models on the SimplerEnv-WidowX benchmark.
}\label{tab:result_simplerenv_widowx}
\vspace{-8pt}
\centering
\resizebox{1.0\columnwidth}{!}{
\begin{small}
\renewcommand{\multirowsetup}{\centering}
\tabcolsep=0.2cm
\renewcommand\arraystretch{1.0}
\begin{tabular}{c|ccccc}
\toprule
\hline
Models & PutSpoon & PutCarrot & StackGreen & PutEggplant & Average \\

\hline
Octo~\cite{team2024octo} &	            47.2 & 8.3 & 4.2 & 56.9 & 29.5 \\
OpenVLA~\cite{kim2024openvla} &	        45.8 & 20.8 & 24.2 & 79.2 & 42.5 \\
SpatialVLA~\cite{qu2025spatialvla} &	16.7 & 25.0 & 29.2 & \textbf{100} & 42.7 \\
CogACT~\cite{li2024cogact} &	        71.1 & 50.8 & 15.0 & 67.5 & 51.3 \\
UniVLA~\cite{wang2025univla} & 	        83.3 & 66.7 & 33.3 & 95.8 & 69.8 \\

\rowcolor{blue!10}
\textbf{VITA (ours)} & \textbf{84.2} & \textbf{68.8} & \textbf{37.5} & 95.6 & \textbf{71.5} \\

\hline
\bottomrule
\end{tabular}
\end{small}
}
\vspace{-15pt}
\end{table}

\subsection{Performance in Real-World}

We evaluate in the real world using a UR-5e~\cite{wu2024robomind} robotic arm. 
As shown in Table~\ref{tab:result_real}, VITA demonstrates significant performances across all six tasks, achieving average success rate of 80.5\%, outperforming all baselines. 
The first four ID tasks indicates VITA's strong adaptability to fundamental perception and decision-making tasks. 
The latter two OOD tasks impose higher demands on abstract reasoning and contextual understanding. Under these challenging conditions, all baseline models exhibit a marked performance drop. For instance, Octo suffers a 48\% decrease in success rate. 
Remarkably, VITA maintains robust performance on these complex scenarios, achieving success rates of 66.9\% and 71.3\%, respectively. 
This highlights VITA's superior task stability and cross-domain generalization capability, establishing a clear advantage in handling unseen, high-difficulty manipulation tasks.
\begin{table}[ht]
\caption{
Real-World Evaluation Results. For each model, we report the average success rate over 1,000 rollouts, where the top four are ID tasks, and bottom two are OOD tasks.
}\label{tab:result_real}
\vspace{-8pt}
\centering
\resizebox{1.0\columnwidth}{!}{
\begin{small}
\renewcommand{\multirowsetup}{\centering}
\tabcolsep=0.1cm
\renewcommand\arraystretch{1.0}
\begin{tabular}{c|ccccc}
\toprule
\hline
\multicolumn{1}{c|}{\multirow{1}{*}{\text{Tasks}}} & 
\textbf{VITA (ours)} & 
Pi0~\cite{black2024pi_0} & 
GR00T N1.5~\cite{bjorck2025gr00tn1.5} & 
OpenVLA~\cite{kim2024openvla} & 
Octo~\cite{team2024octo} \\

\hline
\textbf{Select Object} &        \textbf{92.3} & 60.5 & 47.6 & 49.3 & 52.1 \\
\textbf{Color Match} &          \textbf{87.6} & 56.8 & 49.2 & 61.7 & 24.8 \\
\textbf{Object Map} &           \textbf{91.5} & 58.2 & 52.3 & 46.5 & 48.9 \\
\textbf{Visual Reason} &        \textbf{73.2} & 51.4 & 42.8 & 40.2 & 18.5 \\
\textbf{Inverse Execution} &    \textbf{66.9} & 48.7 & 36.1 & 28.9 & 20.3 \\
\textbf{Conditional Decision} & \textbf{71.3} & 45.3 & 39.4 & 33.6 & 17.2 \\

\hline
\textbf{Average} &              \textbf{80.5} & 53.5 & 44.6 & 43.4 & 29.8 \\

\hline
\bottomrule
\end{tabular}
\end{small}
}
\vspace{-15pt}
\end{table}

\subsection{Effectiveness of Internal CoT}
To investigate the impact of different CoT paradigms on robotic manipulation, we design multiple CoT variants based on same vlm backbone. 
Table~\ref{tab:result_internal} presents the comparison of these variants under identical training and fine-tuning protocols. We draw the following conclusions:
(i) Due to spatial discrepancies across 3D scenes, textual-only CoT exhibits degraded precision in action trajectory control compared to variants that incorporate visual dynamics. 
This highlights the insufficiency of purely symbolic reasoning for grounded motor execution in complex physical environments.
(ii) In contrast to vanilla visual CoT models, internal CoT variants unify perception and action, demonstrating significantly stronger performance on long-horizon, complex manipulation tasks. 
This advantage likely stems from internalizing visual prediction as an inductive bias rather than an explicit decoding target.
\begin{table}[ht]
\vspace{-5pt}
\caption{
Performance comparison of various chain-of-thought strategies in simulated environments.
}\label{tab:result_internal}
\vspace{-8pt}
\centering
\resizebox{1.0\columnwidth}{!}{
\begin{small}
\renewcommand{\multirowsetup}{\centering}
\tabcolsep=0.2cm
\renewcommand\arraystretch{1.0}
\begin{tabular}{c|cccc}
\toprule
\hline
\textbf{CoT Strategy} & \textbf{LIBERO} & \textbf{WidowX} & \textbf{LIBERO-Long} & \textbf{CALVIN(5)} \\

\hline
\textbf{Without CoT} &                    53.7 & 2.5 & 29.8 & 1.83 \\
\textbf{Textual-only CoT} &            	  56.2 & 4.8 & 31.5 & 2.01 \\
\textbf{Visual-only CoT} &	              68.9 & 52.1 & 42.3 & 3.25 \\
\textbf{Textual-Visual CoT} &             72.4 & 63.7 & 45.8 & 3.89 \\
\textbf{Internal CoT} &                   94.1 & 68.3 & 92.7 & 4.52 \\

\rowcolor{blue!10}
\textbf{Textual-Internal CoT (VITA)} &    \textbf{96.7} & \textbf{71.5} & \textbf{96.8} & \textbf{4.67} \\

\hline
\bottomrule
\end{tabular}
\end{small}
}
\vspace{-15pt}
\end{table}

\subsection{Effectiveness of training strategy}
\subsubsection{Setup}
To validate the effectiveness of warmup and co-train strategy, we design the following training variants:
(a) Random all: The shared codebook and both decoder are randomly initialized, which directly to the co-train. We further allow to use two separate codebooks for complete modality decoupling.
(b) + Shared codebook: Codebook trained in warmup is loaded and frozen, while both decoders are randomly initialized.
(c) + Visual decoder: Codebook and visual decoder trained in warmup are loaded.

\begin{table}[ht]
\vspace{-5pt}
\caption{
Performance comparison of various training strategies.
}\label{tab:result_training}
\vspace{-8pt}
\centering
\resizebox{1.0\columnwidth}{!}{
\begin{small}
\renewcommand{\multirowsetup}{\centering}
\tabcolsep=0.2cm
\renewcommand\arraystretch{1.0}
\begin{tabular}{c|cccc}
\toprule
\hline
\textbf{Training Strategy} & \textbf{LIBERO} & \textbf{WidowX} & \textbf{LIBERO-Long} & \textbf{CALVIN(5)} \\

\hline
\textbf{w/o warmup} &              45.2 & 0.0 & 15.9 & 1.29 \\
\textbf{+shared codebook} &       68.7 & 32.1 & 48.5 & 2.87 \\
\textbf{+visual decoder} &	       85.3 & 58.9 & 76.2 & 3.91 \\

\rowcolor{blue!10}
\textbf{+Action decoder (VITA)} & \textbf{96.7} & \textbf{71.5} & \textbf{96.8} & \textbf{4.67} \\

\hline
\bottomrule
\end{tabular}
\end{small}
}
\vspace{-15pt}
\end{table}
\subsubsection{Results and analysis}
Table~\ref{tab:result_training} presents the performance comparison of these ablation variants under identical training protocols. We draw the following conclusions:
(i) Compared to variant (c), variant (b) fails to effectively capture the fine-grained spatial relationships required for precise action execution, resulting in poor performance on high-precision control tasks such as those in the WidowX setting;
(ii) All partial initialization variants (a, b, c) exhibit significant performance gaps compared to the full VITA model, particularly on long-horizon, multi-step tasks such as LIBERO-Long and CALVIN. This is likely because VITA internalizes visual perception and action generation into a unified reasoning process through end-to-end joint optimization.

\subsection{In-Depth Analysis}
\subsubsection{Data and Training Efficiency}
We evaluate model data efficiency in Table~\ref{tab:result_efficiency}~(a). 
VITA fine-tuned on only 10\% of the data surpasses OpenVLA fine-tuned on the full dataset, demonstrating the significant few-shot performance.
Table~\ref{tab:result_efficiency}~(b) presents model training efficiency, showing that VITA achieves convergence in fewer fine-tuning iterations. 
\begin{table}[ht]
\vspace{-5pt}
\caption{
Further experiments in simulated environment to evaluate model data and training efficiency.
}\label{tab:result_efficiency}
\vspace{-8pt}
\centering
\resizebox{1.0\columnwidth}{!}{
\begin{small}
\renewcommand{\multirowsetup}{\centering}
\tabcolsep=0.5cm
\renewcommand\arraystretch{1.0}
\begin{tabular}{cccc}
\toprule
\hline

\multicolumn{4}{c}{\textit{(a) Data efficiency}} \\
\hline

\textbf{Models} & \textbf{Few-shot} & \textbf{CALVIN} & \textbf{Degradation} \\
\hline

\textbf{OpenVLA~\cite{kim2024openvla}} &    10\% & 1.97 & 39.8\% \\
\textbf{UniVLA~\cite{wang2025univla}} &     10\% & 3.19 & 27.6\% \\
\textbf{VITA (w/o warmup)} &                10\% & 0.68 & 47.3\% \\
\rowcolor{blue!10}
\textbf{VITA (ours)} &                      10\% & 3.83 & 17.9\% \\
\hline

\multicolumn{4}{c}{\textit{(b) Training efficiency on CALVIN}} \\
\hline

\textbf{Finetuning Iters.} & \textbf{0.5k} & \textbf{1k} & \textbf{2k} \\
\hline

\textbf{VITA (w/o warmup)} &                0.60 & 0.85 & 1.14 \\
\rowcolor{blue!10}
\textbf{VITA (ours)} &                      4.09 & 4.18 & 4.35 \\

\hline
\bottomrule
\end{tabular}
\end{small}
}
\vspace{-5pt}
\end{table}

\subsubsection{Inference Latency}
In Table~\ref{tab:result_latency}, we report the model's inference latency on the CALVIN benchmark. 
Compared to the ``predict-then-act'' paradigm CoT-VLA~\cite{zhao2025cotvla}, VITA exhibits lower time latency in long-horizon task reasoning. 
Specifically, on an NVIDIA RTX 4090 GPU (24GB), the system achieves an average action-level inference throughput of 60 Hz. 
Details in the Section 10.8 of appendix.
\begin{table}[ht]
\vspace{-5pt}
\caption{
We report the model's inference latency (ms) under different predicted future frame horizons.
}\label{tab:result_latency}
\vspace{-8pt}
\centering
\resizebox{1.0\columnwidth}{!}{
\begin{small}
\renewcommand{\multirowsetup}{\centering}
\tabcolsep=0.7cm
\renewcommand\arraystretch{1.0}
\begin{tabular}{c|ccc}
\toprule
\hline
\textbf{Model} & \textbf{Len=12} & \textbf{Len=24} & \textbf{Len=48} \\

\hline
\textbf{CoT-VLA}	            & 168.2 & 191.1 & 243.9 \\
\textbf{VITA's backbone)}	    & 45.2 & 45.7 & 46.5 \\
\textbf{VITA's action decoder)}	& 15.4 & 18.2 & 24.8 \\

\rowcolor{blue!10}
\textbf{VITA}	                & \textbf{60.6} & \textbf{63.9} & \textbf{71.3} \\

\hline
\bottomrule
\end{tabular}
\end{small}
}
\vspace{-15pt}
\end{table}

\begin{figure}
\begin{center}
\centerline{\includegraphics[width=1.0\columnwidth]{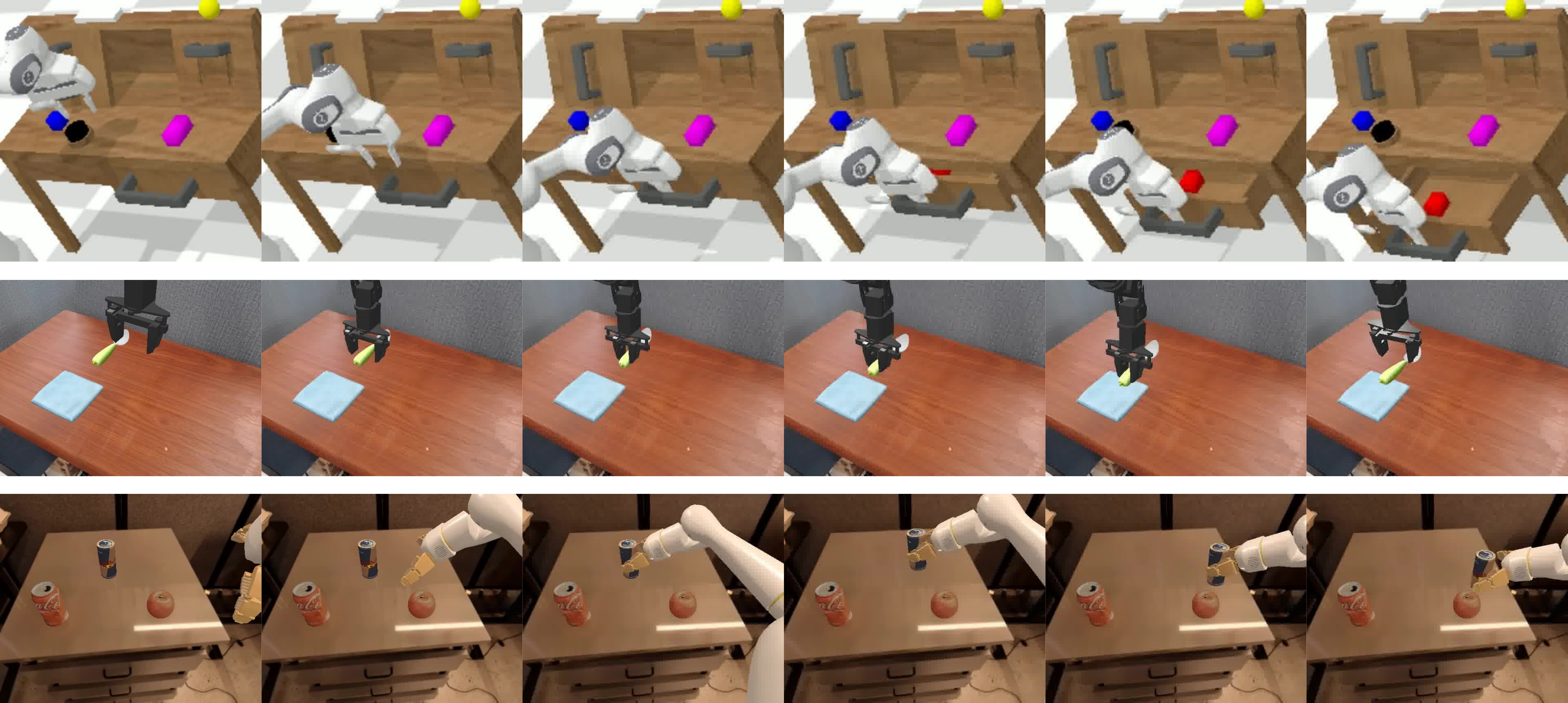}}
\caption{
visualizations of VITA’s generated action trajectories on simulations CALVIN, SimplerEnv Google Robot and WidowX.
}\label{fig:exper4}
\end{center}
\vspace{-30pt}
\end{figure}

\subsection{Visualization}

In Figure~\ref{fig:exper3}, we provide visualizations of VITA’s execution under instruction ``Place red and blue blocks on towels (or plates) of matching colors'' in real-world.
Figure~\ref{fig:exper4} presents visualizations under instruction ``close drawer'' and ``pick up cola can'' on the CALVIN and SimplerEnv benchmark.




\section{Conclusion}\label{section:conclusion}
We propose VITA, a VLA framework that unifies perception and action via a latent space, using future frame prediction as an inductive bias for dynamics. 
Through progressive training on diverse real-world videos, it discards irrelevant pixels and learns generalizable motion. 
VITA achieves +14.5\%, +9.6\%, +12.1\% improvements on CALVIN, LIBERO, and SimplerEnv, and reaches 80.5\% average success across six real-world tasks.

\clearpage
{
    \small
    \bibliographystyle{ieeenat_fullname}
    \bibliography{main}
}

\clearpage
\newpage
\section{The Use of Large Language Models}
During the preparation of this manuscript, we used large language models (LLMs) solely for text polishing and grammatical correction. 
The model design, experimental validation, and analytical reasoning presented in this work were entirely developed and executed by the authors without reliance on LLMs for scientific content generation. 
All research motivation, theoretical formulation, and extensive empirical evaluation are the independent intellectual contributions of the authors.

\section{Implementation Details} 
\subsection{Hybrid Datasets}\label{sec:appendix_hybrid}
Large-scale pre-training on multimodal datasets collected from diverse sources has demonstrated strong generalization capabilities in large language models (LLMs) and vision-language models (VLMs). 
To establish a unified paradigm for task- and cross-embodiment generalist robotic policies, we integrate visual, instructional, and action data from heterogeneous sources. 
Specifically, we construct a complementary multimodal dataset from the following three categories for use in the warmup and fine-tune stages: 
(1) Human demonstration video datasets collected from first-person perspectives, including Something-Something-v2~\cite{goyal2017ssv2} and Ego4D~\cite{grauman2022ego4d}; 
(2) Robot operation videos and action trajectories in real-world environments, including the large-scale open-world manipulation dataset Open X-Embodiment (OXE)~\cite{o2024oxe} and the reasoning-driven robotic benchmark RoboMIND~\cite{wu2024robomind}; 
(3) Robot operation videos and action trajectories in simulated environments, including CALVIN-ABC~\cite{mees2022calvin}.

\begin{table*}
\caption{
We construct a hybrid multimodal dataset for training VITA, comprising: (1) human demonstration video datasets, Something-Something-v2~\cite{goyal2017ssv2} and Ego4D~\cite{grauman2022ego4d}; (2) Robot video-action datasets from real-world settings, Open X-Embodiment (OXE)~\cite{o2024oxe} and RoboMIND~\cite{wu2024robomind}; (3) Robot video-action datasets from simulated environments, CALVIN~\cite{mees2022calvin} and LIBERO~\cite{liu2023libero}.We report the number of videos and total frames in each dataset and compute the relative proportion of each sub-dataset based on frame count.
}\label{tab:appendix_dataset}
\centering
\resizebox{1.5\columnwidth}{!}{
\begin{threeparttable}
\begin{small}
\renewcommand{\multirowsetup}{\centering}
\tabcolsep=0.3cm
\renewcommand\arraystretch{1.4}
\begin{tabular}{ccccc}
\toprule
\hline
\textbf{Datasets} & \textbf{Source} & \textbf{Videos} & \textbf{Frames} & \textbf{Proportion} \\

\hline
\multicolumn{4}{l}{\textit{(1) Human demonstration datasets}} \\
\hline
Something-Something-v2~\cite{goyal2017ssv2} & - & 108499 & 10581365 & 20.47\% \\
Ego4D~\cite{grauman2022ego4d} & - & 59427 & 14202926 & 27.68\% \\

\hline
\multicolumn{4}{l}{\textit{(2) Real-world robot datasets}} \\
\hline
Bridge~\cite{ebert2021bridge1,walke2023bridge2} & OXE~\cite{o2024oxe} & 25460 & 813372 & 1.56\% \\
Fractal~\cite{brohan2022rt1} & OXE~\cite{o2024oxe} & 87212 & 3786400 & 7.41\% \\
FMB Dataset~\cite{luo2025fmb} & OXE~\cite{o2024oxe} & 8611 & 1137340 & 2.14\% \\
Kuka~\cite{kalashnikov2018kuka} & OXE~\cite{o2024oxe} & 209880 & 2455879 & 4.87\% \\
Berkeley Gnm Recon~\cite{shah2021rapid} & OXE~\cite{o2024oxe} & 11834 & 610907 & 1.17\% \\
DROID~\cite{khazatsky2024droid} & OXE~\cite{o2024oxe} & 92233 & 27044326 & 5.26\% \\
Single-Arm Franka & RoboMIND~\cite{wu2024robomind} & 16018 & 2268033 & 4.48\% \\
Single-Arm UR-5e & RoboMIND~\cite{wu2024robomind} & 25721 & 2643322 & 5.07\% \\
AgileX Cobot Magic V2.0 & RoboMIND~\cite{wu2024robomind} & 10059 & 6477564 & 12.67\% \\

\hline
\multicolumn{4}{l}{\textit{(3) Simulation robot datasets}} \\
\hline
CALVIN-ABC~\cite{mees2022calvin} & - & - & 1944237 & 3.70\% \\
LIBERO~\cite{liu2023libero} & - & 6500 & 1865203 & 3.51\% \\

\hline
\bottomrule
\end{tabular}
\end{small}
\end{threeparttable}
}
\end{table*}

These sources encompass a wide range of agent hardware configurations and diverse task scenarios. To train the model to establish a unified motion perception from extensive multimodal data, we address discrepancies in collection frequency across sub-datasets. 
Specifically, to ensure a consistent 1-second time interval between keyframes, we set a dedicated frame sampling stride for each dataset. 
To maintain input consistency, we filter out video sequences lacking high-quality textual annotations and retain only those with more than 12 frames. 
We summarize the distribution of episodes and frames, along with their relative proportions, across these datasets in Table~\ref{tab:appendix_dataset}. 
To appropriately balance the contribution of data collected from different sources and modalities during training, we select specific subsets from OXE~\cite{o2024oxe} and RoboMIND~\cite{wu2024robomind} to form our mixed training set. 
These strategies ensure that the model is consistently exposed to a diverse and balanced mixture of samples in every training iteration. 
This design is crucial for stabilizing optimization and preventing domain over fitting during large-scale pre-training.

\subsection{Warmup and Co-train Details}
To maintain consistency, we resize all camera visual observations to 256×256. 
VITA is configured to predict 12 future frames in parallel during inference, with a maximum sequence length of 8,000 tokens. 
The model is trained on 16 NVIDIA A100 GPUs. 
In the warmup stage, we perform 200K steps of full-parameter training with 324M trainable parameters, requiring approximately 1.5 days. 
In the co-train stage, we conduct 100K steps of full-parameter training with 2.8B trainable parameters, taking about 3.5 days. 
Additional training hyperparameters are provided in Table~\ref{tab:appendix_hyper_parameters}.

\subsection{Finetune Details}
We evaluate VITA in both simulated and real-world environments.
Prior to evaluating performance on simulated benchmarks (e.g., CALVIN~\cite{mees2022calvin}, SimplerEnv~\cite{li2024simplerenv}, and LIBERO~\cite{liu2023libero}) and real-world tasks, we first fine-tune the model parameters. 
During fine-tuning, we load and freeze the pre-trained VLM Backbone visual decoder, while only fine-tuning the action decoder. 
Specifically, for evaluation on the CALVIN benchmark, we fine-tune the model on the CALVIN-ABC dataset and then evaluate its performance on the full CALVIN benchmark. 
Similarly, we fine-tune on the real-world BridgeV2~\cite{walke2023bridge2} dataset and Fractal~\cite{brohan2022rt1} dataset and evaluate on the SimplerEnv benchmark to validate the model’s real-to-sim transfer capability. 
Additionally, we collect over 3,300 real-world robot operation videos paired with action trajectories using a UR-5e robotic arm. The details of the dataset collection protocol can be found in Section~\ref{sec:protocol}.
These real-world datasets are also used to fine-tune both VITA and other baseline models.

Specifically, for evaluation on the CALVIN benchmark~\cite{mees2022calvin}, we fine-tune the model on the CALVIN-ABC dataset using RGB observations from both third-person and robot-mounted (gripper) cameras. 
We set the batch size to 192 and train for 40,000 steps, with each task sequence having a length of 5. 
For the LIBERO benchmark~\cite{liu2023libero}, comprising four task suites, we similarly fine-tune using RGB images from third-person and gripper viewpoints, with a batch size of 192 and 60,000 training steps.

Most notably, the real-to-sim evaluation on SimplerEnv benchmark~\cite{li2024simplerenv} trains policies on real-world data and tests them in simulation. 
The test scenarios are divided into two categories based on data source: Google Robot (Fractal) and WidowX (Bridge V2). 
The Fractal split has a limited domain gap between training and testing, whereas the Bridge V2 branch presents a more challenging transfer scenario. 
For the Fractal benchmark, we use single-view RGB observations, with a batch size of 128 and train for 50,000 steps using an action chunk size of 5. 
For the WidowX (Bridge V2) branch, we extend the training to 80,000 steps under the same configuration. During evaluation, we conduct 24 randomly initialized rollouts across four tasks, Pick Carrot, Pick Eggplant, Pick Spoon, and Stack Cube, and report the final success rate as the performance metric.

Throughout all fine-tuning stages, to maintain consistency with the pre-training process, we uniformly resize all camera visual observations to 256×256.

\subsection{Inference}
Our approach adopts a scheme that simultaneously generates future video frames and aligned action trajectories during training, requiring supervision from both future visual frames and ground-truth actions. 
During inference, VITA retains only the action decoder to produce action tokens, while future frame generation is disabled, enabling efficient and low-latency motor command execution.

\subsection{Multimodal input protocol}
\subsubsection{Input frame}
We follow the multi-frame observation processing protocols of existing methods~\cite{team2024octo,liu2025towards} to ensure fair comparison. 
For GR00T N1.5~\cite{bjorck2025gr00tn1.5}, we concatenate historical observation frames and input them into the three original visual views (left, right, wrist) to enable multi-view modeling. 
For CogACT~\cite{li2024cogact}, since its original VLM supports only single-view, single-timestep processing, we extract the cognitive tokens at each timestep and perform temporal concatenation, thereby providing consistent cross-view semantic constraints for the action expert. 
If a particular view is unavailable in the dataset, the corresponding channel is zero-padded and masked via attention masking to maintain input format consistency.

\subsubsection{State and Action}
For state and action inputs, we construct a unified vector representation capable of accommodating both joint angles and end-effector signals.

\subsubsection{Embodiment}
In single-arm demonstrations, the available arm is mapped to the right-arm channel, while the left-arm channel is zero-padded with masking to maintain compatibility with the dual-arm setup.

\section{Architecture of VITA}\label{sec:appendix_arch}
\subsection{Modules utilized in warmup stage}
To obtain rich semantic representations, we self-supervisedly train a visual decoder in the DINOv2~\cite{oquab2023dinov2} feature space as a forward dynamics model. 
To achieve cross-embodiment generalizable robot action representations, we first normalize multi-source action trajectories and then train an action decoder in the frequency-domain feature space obtained via Discrete Cosine Transform (DCT)~\cite{ahmed2006dct} to model inverse dynamics.

Specifically, the visual encoder is built upon a pretrained DINOv2 encoder and an M-Former~\cite{chen2024moto} module. 
The DINOv2 encoder extracts motion-aware spatiotemporal features from consecutive video frames, and its parameters are frozen during the warmup stage to ensure that training focuses solely on reconstructing dynamic visual content rather than relearning static visual priors. 
In contrast, the visual decoder adopts a Vision Transformer (ViT)~\cite{dosovitskiy2020vit} architecture with 12 transformer layers. 
Notably, we deliberately design this decoder to be lightweight compared to the encoder's capacity. This design reflects the auxiliary role of visual prediction in VLA models: rather than serving as an explicit reconstruction target, it acts as an implicit inductive bias that enhances the model's ability to reason about complex spatial scenes and generate robust motor commands.

Our action encoder consists of a lightweight MLP paired with a Discrete Cosine Transform (DCT), while the action decoder comprises multiple Transformer layers followed by an Inverse DCT. 
The action decoder is deliberately designed with a larger parameter capacity because it bears the core responsibility of accurately reconstructing high-dimensional, continuous, and temporally coherent robot action trajectories from the shared discrete latent space. 
It focuses on capturing fine-grained dynamics from contextual inputs, such as visual observations and task instructions, and ensures that the generated actions are physically smooth and executable.

In contrast, the action encoder only needs to extract compact latent representations, requiring significantly fewer parameters. 
By concentrating model capacity in the decoder, VITA achieves an optimal trade-off: the encoder remains lightweight for efficient training and storage, while the heavy-duty decoder guarantees high precision and robustness in the final motor commands.

During the entire warmup stage, all model parameters are fully trainable except for the DINOv2 encoder, which remains frozen to preserve its rich visual priors and stabilize representation learning.

\subsection{Modules utilized in co-train stage}
To ensure that performance gains stem from modality alignment and the implicit visual chain-of-thought design rather than architectural idiosyncrasies, our VITA implementation follows mainstream design choices established in works like Pi0~\cite{black2024pi_0}. 
Specifically, we initialize the vision-language model (VLM) with the pretrained PaliGemma~\cite{beyer2024paligemma}, which comprises the Gemma~\cite{team2024gemma} language backbone and SigLIP~\cite{zhai2023siglip} as the visual encoder. 
During the co-train and fine-tuning stage, we continue training the VLM backbone, visual decoder, and action decoder on synchronized robot video–action trajectory datasets.

\subsection{Demonstration and reasoning stage}
In demonstration, we retain only the PaliGemma-based backbone, visual decoder, and action decoder. Notably, the visual decoder is used solely for visualization to provide users with interpretable previews of anticipated scene outcomes based on the generated actions. 
However, during actual inference for control, the visual decoder is discarded, and only the action decoder is used to produce executable motor commands, ensuring low-latency operation.

\begin{table}
\caption{
We demonstrated the details of the training hyperparameters and the model architecture.
}\label{tab:appendix_hyper_parameters}
\centering
\resizebox{1.0\columnwidth}{!}{
\begin{threeparttable}
\begin{small}
\renewcommand{\multirowsetup}{\centering}
\tabcolsep=0.3cm
\renewcommand\arraystretch{1.4}
\begin{tabular}{cc}
\toprule
\hline
\textbf{Hyperparameters of pretraining} & \textbf{Value} \\
\hline

Batch size & 1024 \\
Learning rate & 1e-4 \\
Warmup iterations & 200K \\
Co-train iterations & 100K \\
Model precision & Bf16 \\
Optimizer & AdamW \\
Weight decay & 0.01 \\
Optimizer momentum & $\beta_1,\beta_2=0.9,0.95$ \\

\hline
\textbf{Model Architecture} & \textbf{Value/Parameters} \\
\hline
\multicolumn{2}{c}{\textit{Visual Module}} \\
\hline

DINOv2-Base & 86M \\
M-Former & 20M \\
Visual decoder layer & 12 \\
Visual decoder hidden dim & 768 \\
Visual decoder parameter & 96M \\
Visual optimization objective & $\text{L1 loss + SSIM loss}$ \\
Image size & $256\times256$ \\
Predicted future frames & 12 \\

\hline
\multicolumn{2}{c}{\textit{Action Module}} \\
\hline

Action encoder parameter & 10M \\
Action decoder layer & 12 \\
Action decoder hidden dim & 1024 \\
Action decoder parameter & 228M \\
Action optimization objective & $\text{MSE loss}$ \\
Action dimension & 7 \\
Sampling steps & 12 \\

\hline
\multicolumn{2}{c}{\textit{Shared quantizer}} \\
\hline

Size of codebook & 8192 \\
Dimension of encoding vector & 1024 \\

\hline
\multicolumn{2}{c}{\textit{VLM backbone}} \\
\hline

SigLIP (Visual Tokenizer) & 400M \\
Gemma backbone & 2B \\

\hline
Total parameters & 2.8B \\
\hline
\bottomrule
\end{tabular}
\end{small}
\end{threeparttable}
}
\end{table}

\begin{algorithm*}[ht]
\caption{VITA Training and Inference Pipeline}\label{sec:appendix_algo}
\label{alg:vita}
\begin{algorithmic}[1]
\REQUIRE Dataset $D$
\REQUIRE Shared codebook $\mathcal{C} = \{c_k\}_{k=1}^K$, visual encoder $\mathcal{E}_v$, action encoder $\mathcal{E}_a$
\REQUIRE Visual decoder $\mathcal{D}_v$, action decoder $\mathcal{D}_a$

\STATE \textbf{/* Stage 1: Warmup */}
\FOR{each batch from $D$}
    \IF{sample is video pair $(I_t, I_{t+1})$}
        \STATE $z_v \gets \text{M-Former}([DINOv2(I_t); DINOv2(I_{t+1})])$
        \STATE $\hat{z}_v \gets \mathcal{Q}(z_v)$
        \STATE $\hat{I}_{t+1} \gets \mathcal{D}_v(DINOv2(I_t), \hat{z}_v)$
        \STATE $L_v = \lambda_{L1} \|I_{t+1} - \hat{I}_{t+1}\|_1 + \lambda_{SSIM}(1 - SSIM(I_{t+1}, \hat{I}_{t+1}))$
        \STATE Update $\mathcal{E}_v$, $\mathcal{D}_v$, $\mathcal{C}$ via $\nabla L_v$
    \ELSIF{sample is action segment $a_{t:t+H}$}
        \STATE $\hat{a} \gets \text{DCT}(a_{t:t+H})$
        \STATE $z_a \gets \text{MLP}(\text{flatten}(\hat{a}))$
        \STATE $\hat{z}_a \gets \mathcal{Q}(z_a)$
        \STATE $\hat{a}_{t:t+H} \gets \text{MLP}(\text{InverseDCT}(\hat{z}_a))$
        \STATE $L_a = \|a_{t:t+H} - \hat{a}_{t:t+H}\|_2^2$
        \STATE Update $\mathcal{E}_a$, $\mathcal{D}_a$, $\mathcal{C}$ via $\nabla L_a$
    \ENDIF
\ENDFOR

\STATE \textbf{/* Stage 2: Co-train */}
\FOR{each batch from $D$}
    \STATE Input: instruction $x$, image $I_0$, state $s$
    \STATE $h_{\text{ctx}} \gets [\mathcal{T}_{\text{text}}(x); \mathcal{T}_{\text{image}}(I_0); s]$
    \STATE Generate textual CoT: $Z_{\text{sub}} = [z_1, ..., z_M]$ via $\arg\max p_\theta(z \mid h_{\text{ctx}})$
    \STATE $h_{\text{mul}} \gets [h_{\text{ctx}}; \mathcal{T}_{\text{text}}(Z_{\text{sub}})]$
    \STATE Autoregressively sample hybrid tokens: $\{\tau_i\}_{i=1}^L \sim p_\theta(\cdot \mid h_{\text{mul}})$
    \STATE Decode: $\hat{I}_{1:T} \gets \mathcal{D}_v(\{c_{\tau_i}\})$, $\hat{a}_{1:H} \gets \mathcal{D}_a(\{c_{\tau_i}\})$
    \IF{only video available}
        \STATE $L_{\text{co}} \gets \sum_{t=1}^T (\lambda_{L1} \|I_t - \hat{I}_t\|_1 + \lambda_{SSIM}(1 - SSIM(I_t, \hat{I}_t)))$
    \ELSE
        \STATE $L_{\text{co}} \gets \lambda_v \|I_{1:T} - \hat{I}_{1:T}\|_1 + \lambda_a \|a_{1:H} - \hat{a}_{1:H}\|_2^2$
    \ENDIF
    \STATE Update $p_\theta$, $\mathcal{D}_v$, $\mathcal{D}_a$ (freeze $\mathcal{C}$)
\ENDFOR

\STATE \textbf{/* Stage 3: Fine-tune */}
\FOR{each batch from task-specific data}
    \STATE Freeze $p_\theta$ and $\mathcal{C}$
    \STATE Update only $\mathcal{D}_a$ using $\|a_{1:H} - \hat{a}_{1:H}\|_2^2$
\ENDFOR

\STATE \textbf{/* Inference */}
\REQUIRE $x$, $I_0$, $s$
\STATE Generate $\{\tau_i\}_{i=1}^L \sim p_\theta(\cdot \mid h_{\text{mul}})$ as in Co-train
\STATE $\hat{a}_{1:H} \gets \mathcal{D}_a(\{c_{\tau_i}\})$
\RETURN $\hat{a}_{1:H}$
\end{algorithmic}
\end{algorithm*}
\subsection{Model architecture details and training hyperparameters}
To facilitate reproducibility, we detail the architecture and training hyperparameters of all VITA components in Table~\ref{tab:appendix_hyper_parameters}. 
Specifically, The SigLIP-based visual encoder contains 400 million parameters. 
The Gemma Transformer backbone has approximately 2 billion parameters. Built upon these strong vision–language priors, we design a 96 million visual decoder and a 228 million action decoder to effectively capture visual dynamics and motion representations, respectively. 
Furthermore, we provide the pseudocode for the warmup, co-train and fine-tune stage in Algorithm~\ref{sec:appendix_algo}.

\section{Details of the Simulation Environment}\label{sec:appendix_simulation}
\subsection{CALVIN}
The CALVIN simulation benchmark is designed for learning long-horizon robotic control policies and comprises four distinct manipulation environments. 
In each environment, multimodal inputs, including language instructions, multi-view RGB-D cameras, tactile, and proprioceptive sensors, are provided. 
CALVIN supports 34 manipulation tasks and 1,000 language instructions. 
In the most challenging ABC-D setup, models are first trained in the ABC environments, followed by evaluation on task completion in environment D, assessing the model's capability for long-horizon planning and zero-shot generalization.

\subsection{LIBERO}
The LIBERO benchmark comprises four long-horizon robotic manipulation test suites, each containing 10 tasks with 50 human demonstrations per task. These suites emphasize distinct aspects of robot control: 
(1) LIBERO-Goal evaluates goal-conditioned behavior by varying task objectives; 
(2) LIBERO-Spatial assesses spatial reasoning through changes in object layout while keeping the target objects fixed; 
(3) LIBERO-Object tests object-level generalization by altering object instances within a fixed scene configuration; 
(4) LIBERO-Long features long-horizon, compositional tasks that require complex sequential planning and execution.

\subsection{SimplerEnv}
To address the limitations of simulated environments in terms of diversity in lighting, color, texture, and robot camera viewpoints compared to real-world scenarios, SimplerEnv integrates multiple robot configurations. 
Specifically, to compare with state-of-the-art generalist manipulation policies, we evaluate VITA's performance under two settings: Google Robot and WidowX.

\section{Real-world robot platform}
\subsection{UR-5e Experimental Setup}
\begin{figure}[t]
\begin{center}
\centerline{\includegraphics[width=1.0\columnwidth]{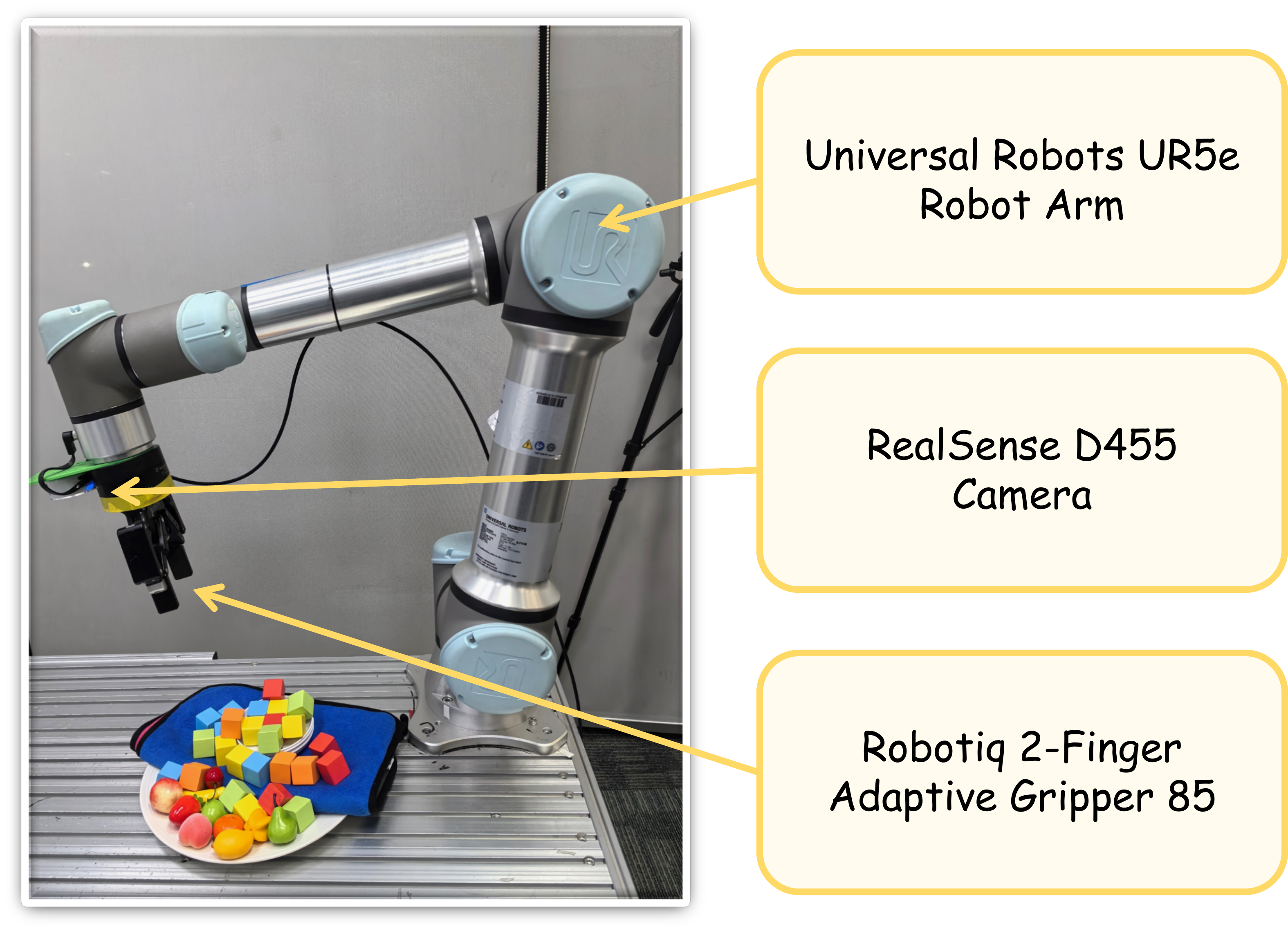}}
\caption{
Demonstration of our real-world robotic platform.
}\label{fig:appendix_ur5e}
\end{center}
\vspace{-20pt}
\end{figure}

\begin{figure*}[t]
\begin{center}
\centerline{\includegraphics[width=2.0\columnwidth]{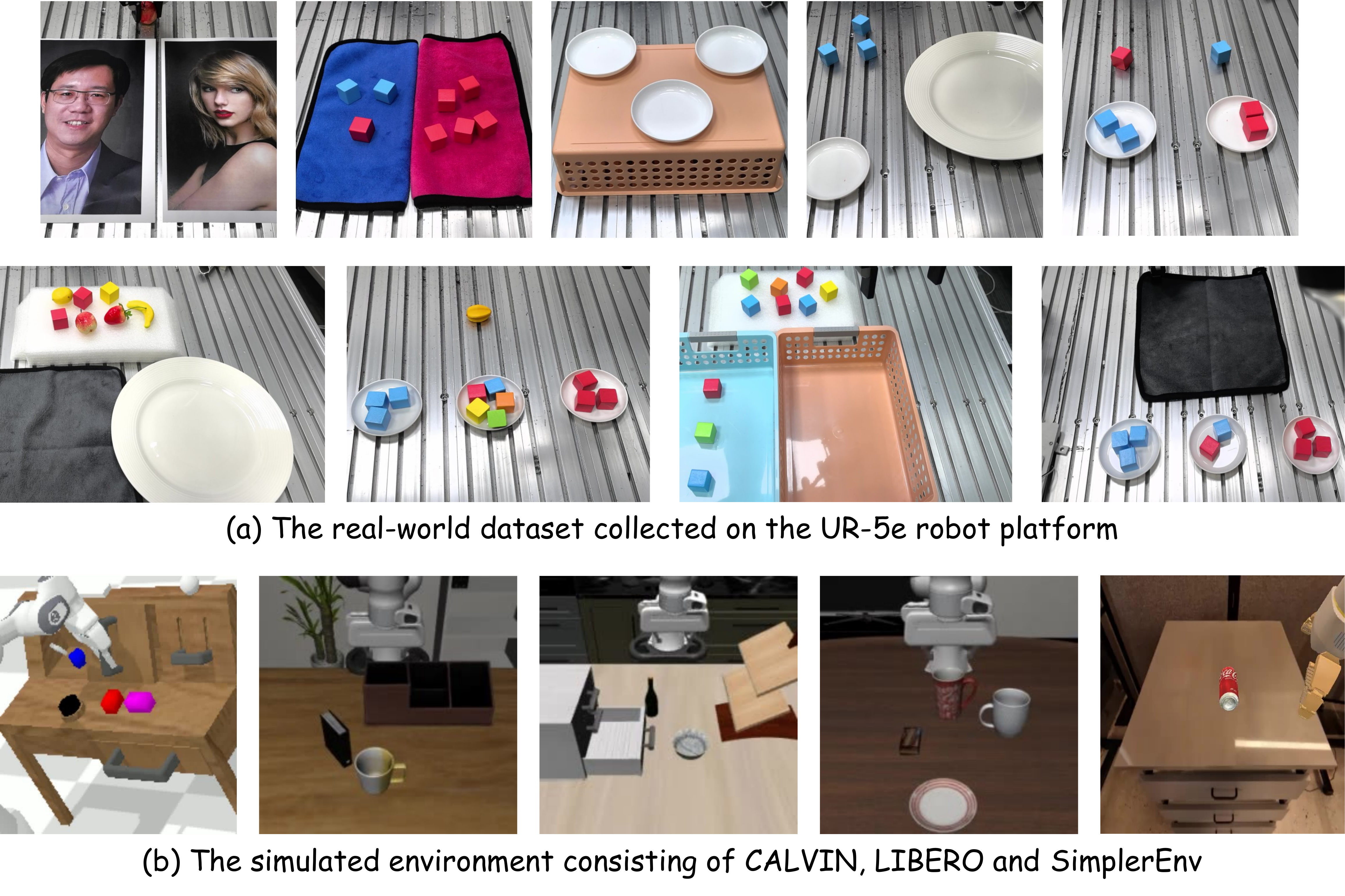}}
\caption{
(a) Based on the established real-world robotic platform, we designed multiple scenarios to collect training data for the UR-5e agent. (b) Furthermore, we also presented the benchmark in the simulation environment.
}\label{fig:appendix_ur5e_scenes}
\end{center}
\vspace{-20pt}
\end{figure*}

As shown in Figure~\ref{fig:appendix_ur5e}, we use a Universal Robots UR5e (UR-5e) robotic arm equipped with a Robotiq 2-Finger Adaptive Gripper 85. 
The robot is configured with two cameras providing distinct viewpoints: a first-person view (camera mounted above the end-effector) and a third-person view (camera fixed at the edge of the table). 
Additionally, we present a series of manipulation tasks collected from real-world scenarios in Figure~\ref{fig:appendix_ur5e_scenes}.

\begin{table*}[ht]
\caption{
Real-World task trajectories collected from the Robots UR-5e. In the second column of the table, we use numbers to represent the corresponding training tasks. The complete definitions of the 12 tasks can be found in Section~\ref{sec:task_definition}.
}\label{tab:appendix_ur5e}
\vspace{-8pt}
\centering
\resizebox{2.0\columnwidth}{!}{
\begin{small}
\renewcommand{\multirowsetup}{\centering}
\tabcolsep=0.3cm
\renewcommand\arraystretch{1.0}
\begin{tabular}{cccccc}
\toprule
\hline
\textbf{Task type} & \textbf{Task ID} & \textbf{Scenes} & \textbf{Frames/Scene} & \textbf{Frames} & \textbf{Proportion} \\
\hline

Basic Manipulation and Short-term Reasoning & Task$_{1-5}$ & 300 & 209 & 314,592 & 27.1\% \\
Multi-step Execution & Task$_{6-8}$ & 200 & 473 & 283,607 & 24.4\% \\
Textual-Visual Reasoning & Task$_{9-10}$ & 400 & 321 & 256,854 & 22.1\% \\
Long-horizon Planning & Task$_{11-12}$ & 200 & 768 & 307,364 & 26.4\% \\

\rowcolor{blue!10}
UR-5e Single Arm Dataset (Total) & - & 3,300 & 352 & 1,162,417 & 100\% \\

\hline
\bottomrule
\end{tabular}
\end{small}
}
\vspace{-8pt}
\end{table*}
\subsection{Real-World Train Dataset Protocol}\label{sec:protocol}
We summarize the real-world data collected from the UR-5e robotic platform in Table~\ref{tab:appendix_ur5e}. 
All robot control commands are executed at an actual frequency of 30 Hz. We designed a total of 12 tasks, categorized into four types, covering a full spectrum of capabilities from basic manipulation to long-horizon reasoning. 
We have clearly defined these 12 task scenarios, including execution steps, success criteria, failure conditions, and instruction templates. Details can be found in Section~\ref{sec:task_definition}.
Each task includes 200–800 randomized scenes to ensure generalization. 
For each task, we collect 3,300 robot action trajectories paired with synchronized operation videos by varying the initial positions of objects on the tabletop, forming the UR-5e dataset.

Our randomization strategy is as follows: 
(1) In each scene, object initial positions are uniformly sampled within the 0.8m×0.8m workspace to avoid fixed patterns; 
(2) Towels and plates are randomly offset by ±15 cm within each scene; 
(3) While instruction templates remain consistent across different instances of the same task, object combinations, quantities, and color distributions are randomly varied to enhance diversity and test robustness.

In our data collection protocol, we operate at the granularity of the robot's minimal executable action: for every teleoperated command issued, we synchronously capture one RGB frame and record the full robot state, including joint angles, end-effector pose, and gripper width. 
The resulting dataset comprises 1,162,417 time steps, each strictly aligned with one robot action command and one 640×480 RGB image captured by a RealSense D455 camera, along with the complete proprioceptive state at that moment.

During preprocessing, each RGB image is resized to 256×256 (uint8) to match the model input specification, and the corresponding depth map is also downsampled to 256×256 (float32). 
Each robot action is represented as a 7-dimensional float32 vector. 
Additionally, we retain comprehensive metadata for every time step, including: full robot state, natural language instruction (average length 50 characters), timestamp, and camera calibration parameters. 

\subsection{Real-World Task Definition}\label{sec:task_definition}
In this section, we provide detailed designs for each task, including execution steps, success criteria, failure conditions, and language instruction templates, to facilitate data annotation and model evaluation.

\vspace{8pt}
\noindent
\textbf{Task 1: Select Celebrity Portrait} \\
The robot must identify a specified celebrity portrait (e.g., ``Taylor Swift'') from a set of displayed images. 
It then moves its gripper to a position 5 cm above the target portrait and slowly lowers it until it makes light contact with the surface, applying a controlled force of no more than 2 N. 
Success is defined as stable contact with the correct portrait without slippage or tipping. Failure occurs if the gripper touches an incorrect portrait, applies excessive force, or fails to make any contact. 
The instruction template includes: ``Please gently touch Taylor Swift's portrait.'' and ``Point to Anne Hathaway's photo with your finger.''

\vspace{8pt}
\noindent
\textbf{Task 2: Color Match to Towel} \\
The robot identifies two colored blocks (red and blue) and two corresponding colored towels. 
It must place the red block entirely on the red towel and the blue block on the blue towel. Success requires both blocks to be correctly positioned with no misplacement. 
Failure occurs if blocks are placed on the wrong towel, fall off during placement, or cause the towel to be pulled or deformed. Instruction templates include: ``Place the red block on the red towel and the blue block on the blue towel.'' and ``Sort by color: red to red towel, blue to blue towel.''

\vspace{8pt}
\noindent
\textbf{Task 3: Color Match to Plate} \\
Identical to Task 2, but the target containers are plates instead of towels. 
The robot must place the red block inside the red plate and the blue block inside the blue plate. 
Success is achieved when both blocks are fully contained within their respective plates, with no part of the block extending beyond the rim. 
Failure occurs if blocks roll out of the plates or are placed on the wrong plate. 
Instruction templates include: ``Put the red block into the red plate and the blue block into the blue plate.'' and `Sort the blocks by color into the corresponding plates.''

\vspace{8pt}
\noindent
\textbf{Task 4: Fold Towel} \\
The robot grasps one end of a flat towel and folds it once along its central axis, aligning the edges as precisely as possible. 
Success is defined as a neat, rectangular fold with minimal wrinkles or distortion. 
Failure occurs if the fold is uneven, the towel slips from the gripper, or multiple folds are attempted. 
The instruction template is: ``Please fold this towel once and make it neat.''

\vspace{8pt}
\noindent
\textbf{Task 5: Stack Cups/Bowls} \\
The robot identifies a stack of paper cups or bowls and sequentially lifts and places each item on top of the previous one to form a stable vertical tower. 
Success requires the final stack to remain upright and intact without any collapse. 
Failure occurs if the stack topples during or after placement. 
The instruction template is: ``Please stack the paper cups one by one.''

\vspace{8pt}
\noindent
\textbf{Task 6: Error Correction (Color Matching)} \\
The robot observes a scene where some blocks are incorrectly placed (e.g., three blue blocks and one red block on a blue towel). 
It must detect the anomaly (the red block on the blue towel), retrieve it, and relocate it to the correct towel (red towel). 
Success is achieved when all blocks are correctly matched to their respective colors with no omissions or new errors. 
Failure occurs if the robot fails to detect the error, moves a correct block, or creates a new misplacement. 
Instruction templates include: ``Check if any blocks are misplaced? If so, correct them.'' and ``There's a red block on the blue towel, which is wrong. Please move it to the red towel.''

\vspace{8pt}
\noindent
\textbf{Task 7: Wipe Plate (Anti-Tip Test)} \\
A plate is balanced precariously atop three stacked blocks, creating an unstable platform. 
The robot must use a gray towel to gently wipe the surface of the plate without disturbing the blocks beneath. 
Success is defined as the plate remaining upright and the towel completing a full wiping motion. Failure occurs if the plate tips over, the blocks are displaced, or the towel fails to contact the plate surface. 
Instruction templates include: ``Gently wipe this plate with the gray towel, be careful not to touch the blocks underneath.'' and ``Be cautious, the plate is wobbly, don’t let it fall.''

\vspace{8pt}
\noindent
\textbf{Task 8: Multi-step Transfer (Large Plate $\rightarrow$ Small Plate)} \\
The robot must perform a two-step sequence: First, transfer all blue blocks from the table into a large plate and wait for 5 seconds (simulating user confirmation). 
Then, transfer all blue blocks from the large plate into a smaller plate. Success requires both steps to be completed in sequence without interruption and with no blocks lost or dropped. 
Failure occurs if the robot begins the second step before the first is complete, or if blocks fall during either transfer. Instruction templates include: ``First, put all blue blocks into the large plate, then transfer them all to the small plate.'' and ``Step 1: Blue blocks into large plate; Step 2: Blue blocks into small plate.''

\vspace{8pt}
\noindent
\textbf{Task 9: Return to Original Position (Towel $\rightarrow$ Plate $\rightarrow$ Towel)} \\
The robot begins with red and blue blocks correctly placed on their respective colored towels. 
It must first collect all blocks and place them into a central plate, then return each block to its original towel. 
Success is defined as the final configuration matching the initial state exactly. 
Failure occurs if blocks are returned to the wrong towel or if any block is missing. 
Instruction templates include: ``First, collect all blocks into the plate, then return them to their original positions.'' and ``Remember where they started; they must return there at the end.''

\vspace{8pt}
\noindent
\textbf{Task 10: Imitate Arrangement Strategy} \\
The robot observes a reference arrangement in a blue basket, e.g., a stack of blocks from top to bottom: red, green, blue. 
It must then replicate this exact vertical order by arranging scattered blocks of the same colors into a red basket. 
Success is achieved when the sequence of colors in the red basket matches the reference sequence precisely. 
Failure occurs if the order is altered, inverted, or incomplete. 
The instruction template is: ``Look at how the blocks are arranged in the blue basket; replicate that arrangement with the blocks on the table in the red basket.''

\vspace{8pt}
\noindent
\textbf{Task 11: Categorize by Object Type (Fruit/Blocks)} \\
The robot identifies objects as either fruit (e.g., apple, banana) or blocks (e.g., colored cubes). 
It must place all fruits into a plate and all blocks onto a gray towel. 
Success requires accurate classification and correct placement with no misplacement. 
Failure occurs if fruits are placed on the towel or blocks are placed in the plate. 
Instruction templates include: ``Put the fruits in the plate and the blocks on the gray towel.'' and ``Distinguish between fruits and blocks: fruits go in the plate, blocks go on the towel.''

\vspace{8pt}
\noindent
\textbf{Task 12: Categorize by Object Color (Red/Green)} \\
The robot identifies red and green objects, including both fruits and blocks, and must place all red items into the plate and all green items onto the gray towel. 
Success is achieved through accurate color-based categorization with no misplacements or omissions. 
Failure occurs if red objects are placed on the towel, green objects are placed in the plate, or any object is left unsorted. 
Instruction templates include: ``Put red items in the plate and green items on the towel.'' and ``Regardless of type, put anything red in the plate and anything green on the towel.''

\subsection{Real-Robot Finetuning.}
For practical evaluation, we conducted real-world experiments on the UR-5e robotic platform using images captured from two camera perspectives: one fixed at the edge of the tabletop (third-person view) and one mounted on the robotic arm (first-person view). 
All input images were resized to a resolution of 256×256. The model outputs a 7-dimensional action vector, with an action chunk size of 15. For each task, we used a learning rate of $5\times10^{-5}$, a batch size of 128, and trained for 40,000 steps.

\subsection{Evaluation on the Real-World Experiment}
We designed a series of representative and challenging real-world tasks to thoroughly evaluate VITA's performance in robotic motion planning under both in-distribution (ID) and out-of-distribution (OOD) settings. 
ID tasks are those that appear in the training set, whereas OOD tasks are held out during training and are specifically included to assess the model's generalization capability. 
These tasks span a broad spectrum of real-world scenarios, including dexterous manipulation, long-horizon action sequences, visual-contextual reasoning, and dynamic interaction with the environment. 
For each task, we report the average success rate over 100 repeated trials. Collectively, these experiments demonstrate VITA's robustness and generalization across diverse real-world conditions. The 4 ID tasks and 4 OOD tasks we designed are as follows:

\vspace{8pt}
\noindent
\textbf{Evaluation 1: Select Object (ID)} \\
Select the Kaiming He's portrait and place a fruit on it.

\vspace{8pt}
\noindent
\textbf{Evaluation 2: Color Match (ID)} \\
Place the red and blue blocks onto the towel of their corresponding colors.

\vspace{8pt}
\noindent
\textbf{Evaluation 3: Object Map (ID)} \\
Put red objects on the towel, and yellow objects in the plate.

\vspace{8pt}
\noindent
\textbf{Evaluation 4: Visual Reason (ID)} \\
Put the fruit into the plate that contains the most blue blocks.

\vspace{8pt}
\noindent
\textbf{Evaluation 5: Inverse Execution (OOD)} \\
First move the blocks from the towel into the plate, then return them to their original positions.

\vspace{8pt}
\noindent
\textbf{Evaluation 6: Conditional Decision (OOD)} \\
If the letter in the plate is "B", place a strawberry on plate.

\subsection{Visualization of Real-World Task}
\begin{figure*}[t]
\begin{center}
\centerline{\includegraphics[width=2.0\columnwidth]{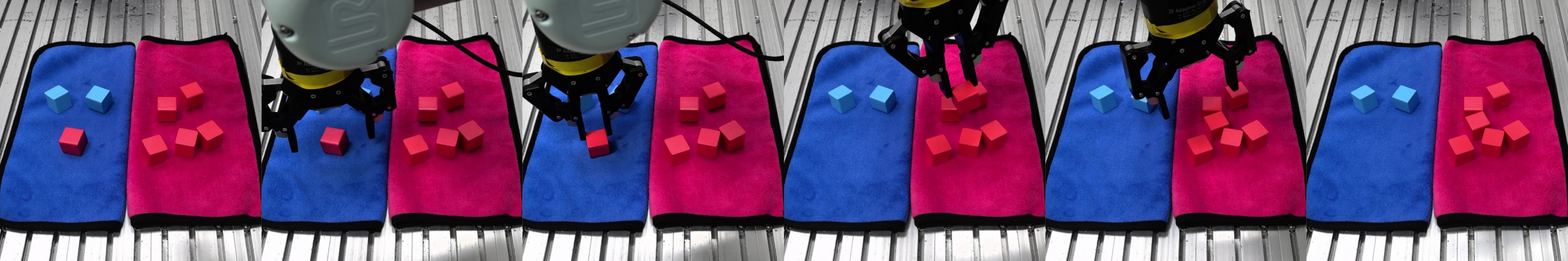}}
\caption{
VITA executes the ``Color Matching'' task. Following the instruction (e.g., ``Place red and blue blocks on towels of matching colors''), VITA controls the gripper to place the red block on the red towel and the blue block on the blue towel, successfully completing the classification.
}\label{fig:appendix_task2}
\end{center}
\vspace{-20pt}
\end{figure*}

\begin{figure*}[t]
\begin{center}
\centerline{\includegraphics[width=2.0\columnwidth]{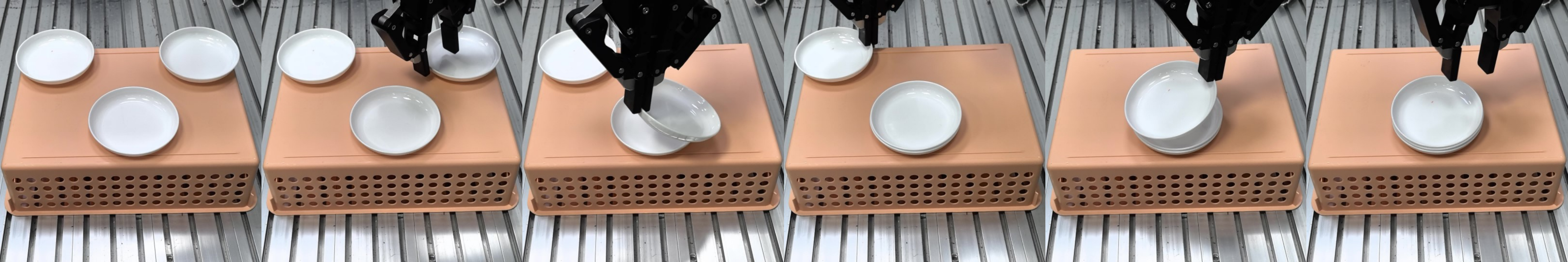}}
\caption{
VITA executes the ``Stack Plates'' task.
Following the instruction (e.g., ``Stack the plates one by one''), VITA controls the gripper to stack them sequentially. 
This process demonstrates VITA's planning and execution capabilities in multi-step manipulation and physical interaction.
}\label{fig:appendix_task3}
\end{center}
\vspace{-20pt}
\end{figure*}

\begin{figure*}[t]
\begin{center}
\centerline{\includegraphics[width=2.0\columnwidth]{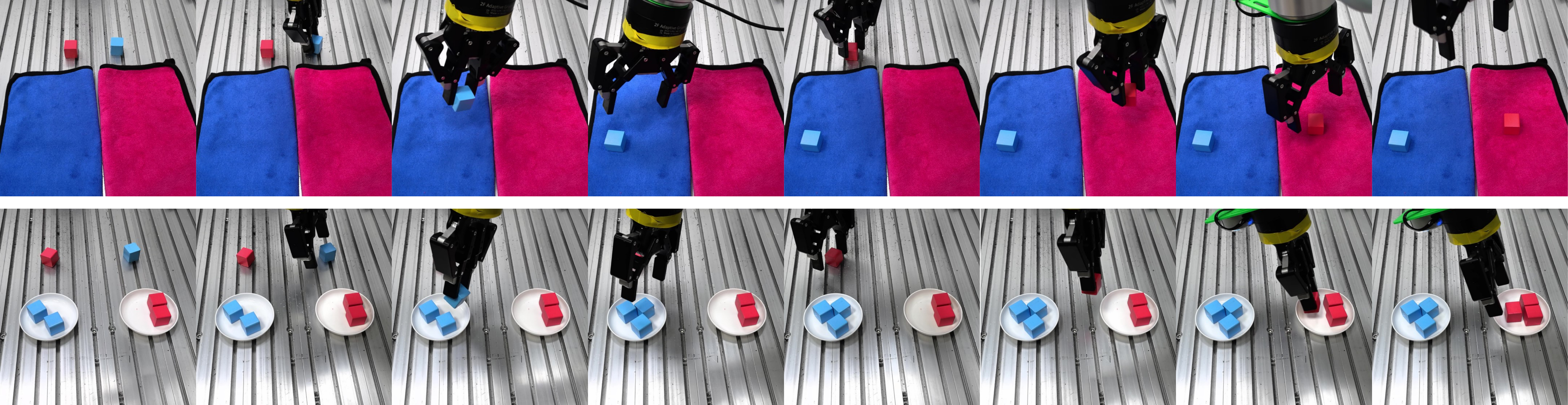}}
\caption{
VITA executes the ``Color Matching'' task.
Following the instruction (e.g., ``Place red and blue blocks on towels (or plates) of matching colors''), VITA controls the gripper to place the blue block on the blue towel (plate) and the red block on the red towel (plate).
}\label{fig:appendix_task4}
\end{center}
\vspace{-20pt}
\end{figure*}

\begin{figure*}[t]
\begin{center}
\centerline{\includegraphics[width=2.0\columnwidth]{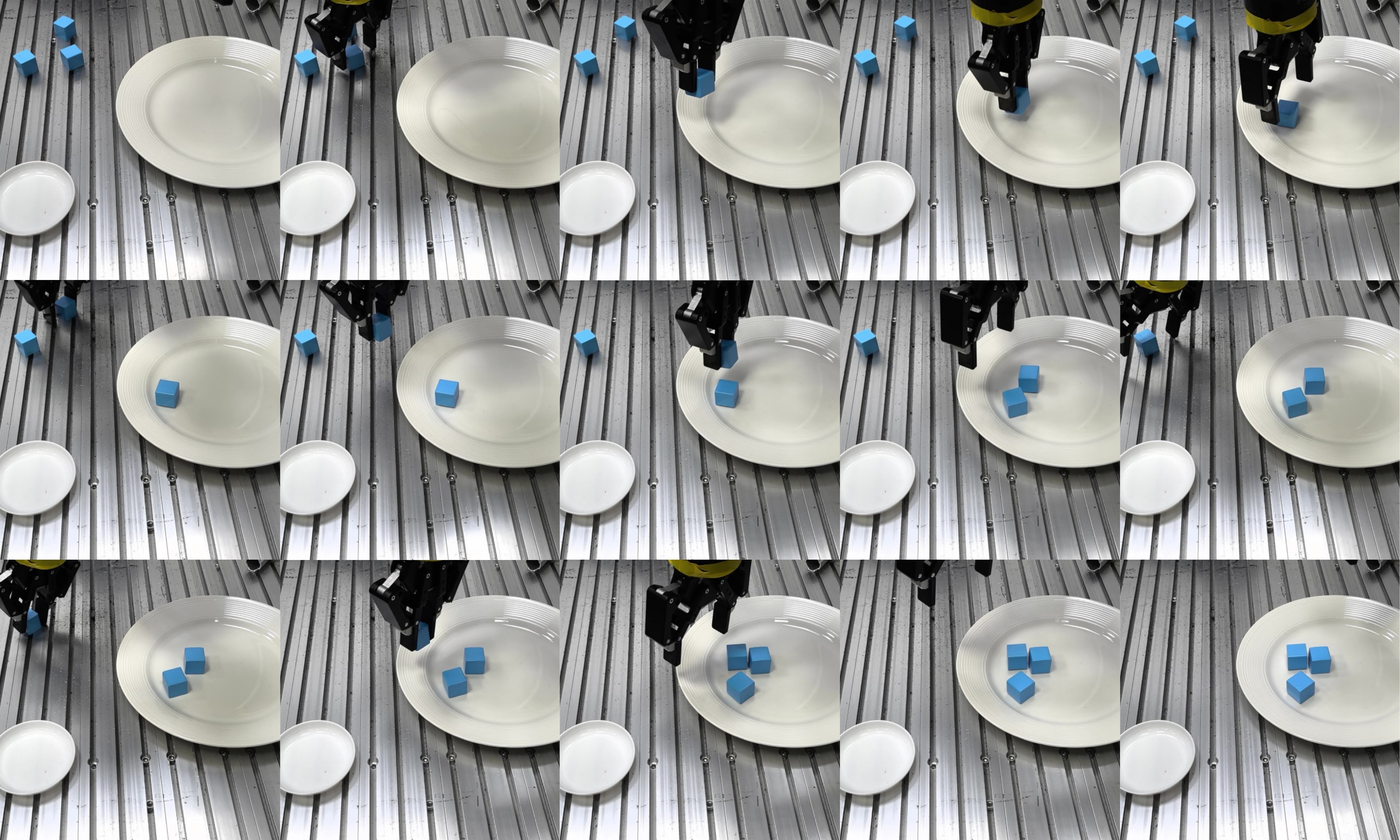}}
\caption{
VITA executes the ``Multi-step Transfer'' task.
Following the instruction (e.g., ``Move all blue blocks from the table to the large plate''), VITA first picks up each blue block scattered on the table and places them into the large plate.
}\label{fig:appendix_task5}
\end{center}
\vspace{-20pt}
\end{figure*}

\begin{figure*}[t]
\begin{center}
\centerline{\includegraphics[width=2.0\columnwidth]{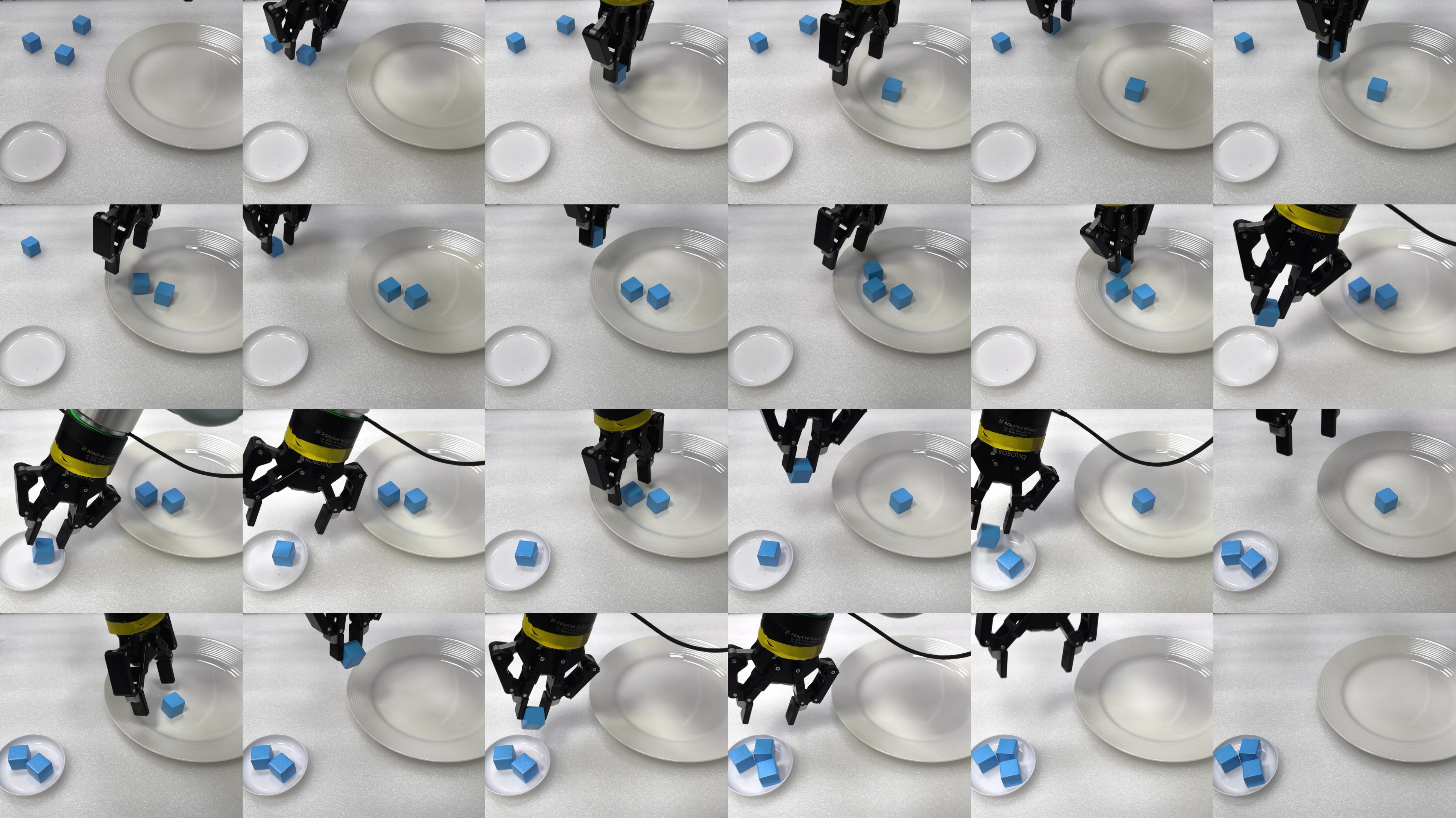}}
\caption{
VITA executes the ``Long-horizon'' task.
Following the instruction (e.g., ``Move all blue blocks from the large plate to the small plate''), VITA first picks up all blue blocks scattered on the table and places them into the large plate, then retrieves them and neatly stacks them into the small plate.
}\label{fig:appendix_task6}
\end{center}
\vspace{-20pt}
\end{figure*}

\begin{figure*}[t]
\begin{center}
\centerline{\includegraphics[width=2.0\columnwidth]{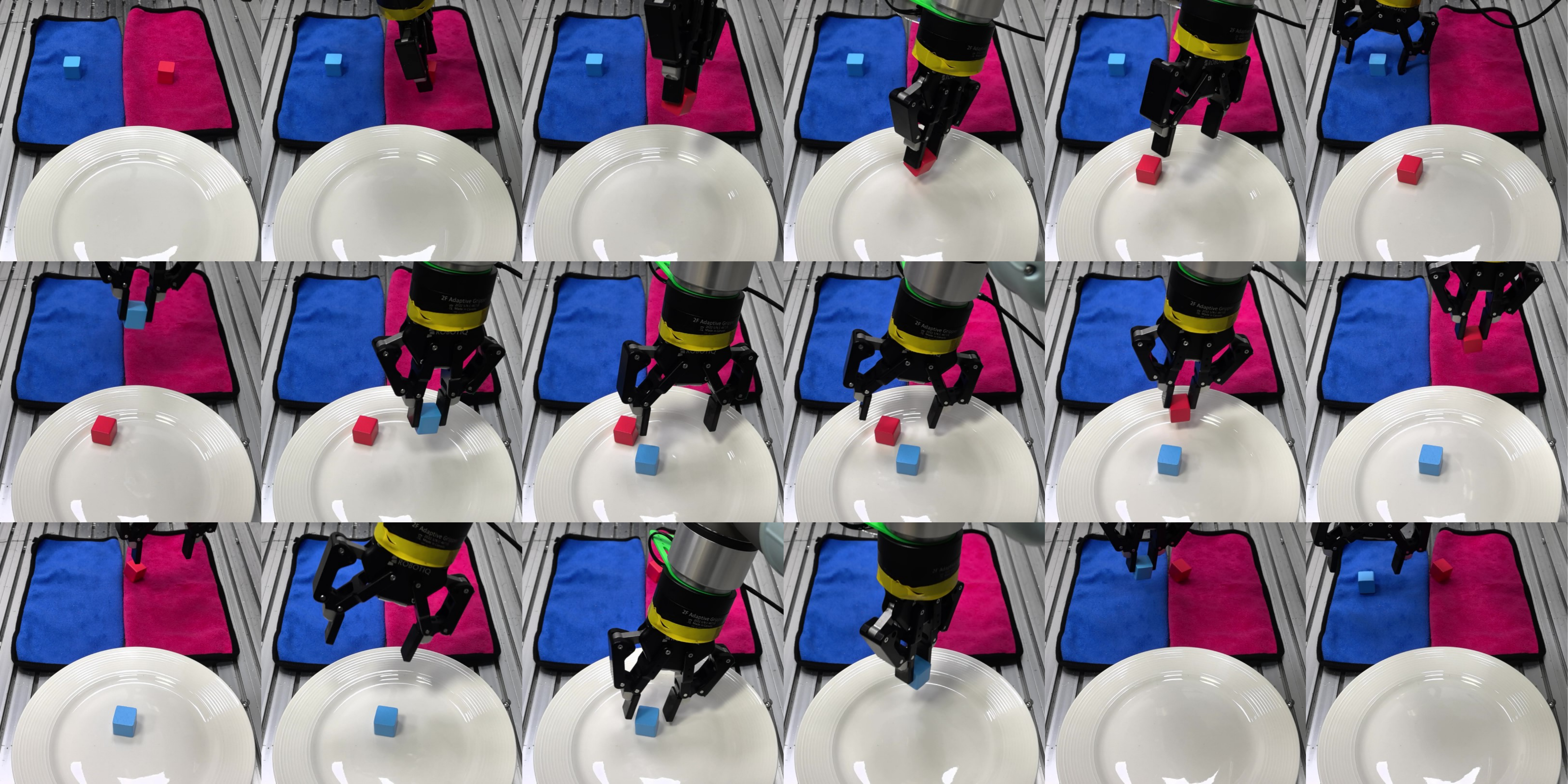}}
\caption{
VITA executes the ``Multi-step Transfer and Restore'' task.
Following the instruction (e.g., ``Move all blocks from the towel to the plate, then restore them to their original positions''), this task evaluates the model's long-horizon planning and contextual memory. 
It requires the model not only to perform simple pick-and-place actions but also to understand and execute a two-stage composite instruction: first moving objects from their initial position (towel) to an intermediate location (plate), then using memory to return them precisely to their original positions.
}\label{fig:appendix_task7}
\end{center}
\vspace{-20pt}
\end{figure*}

\begin{figure*}[t]
\begin{center}
\centerline{\includegraphics[width=2.0\columnwidth]{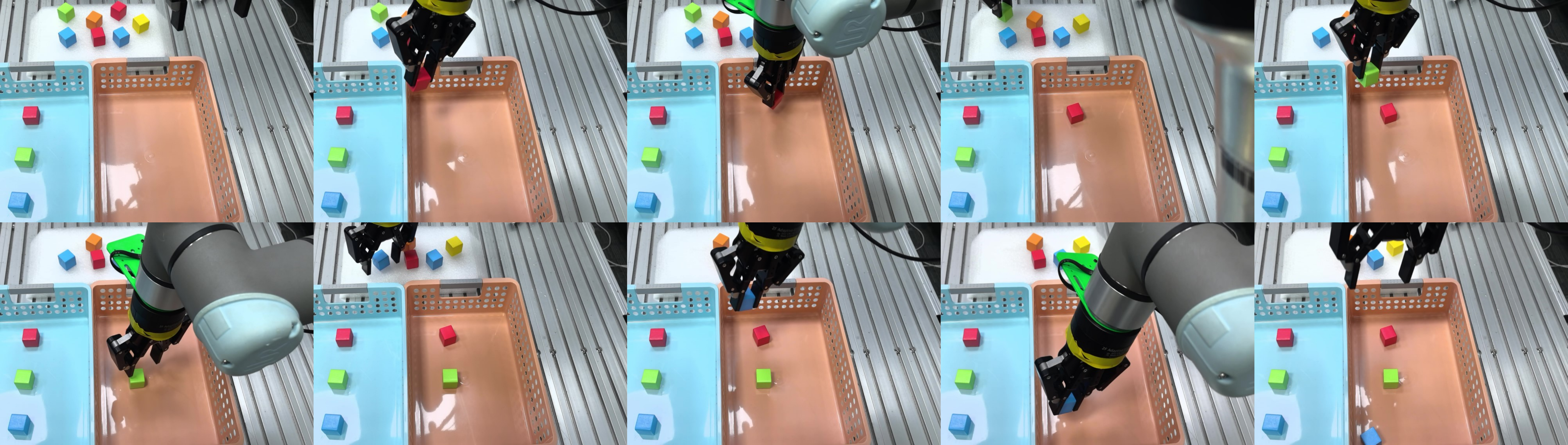}}
\caption{
VITA executes the ``Imitate Arrangement Strategy'' task.
Following the instruction (e.g., ``Arrange scattered blocks in the red basket following the same pattern''), VITA replicates the observed spatial sequence.
}\label{fig:appendix_task8}
\end{center}
\vspace{-20pt}
\end{figure*}

\begin{figure*}[t]
\begin{center}
\centerline{\includegraphics[width=2.0\columnwidth]{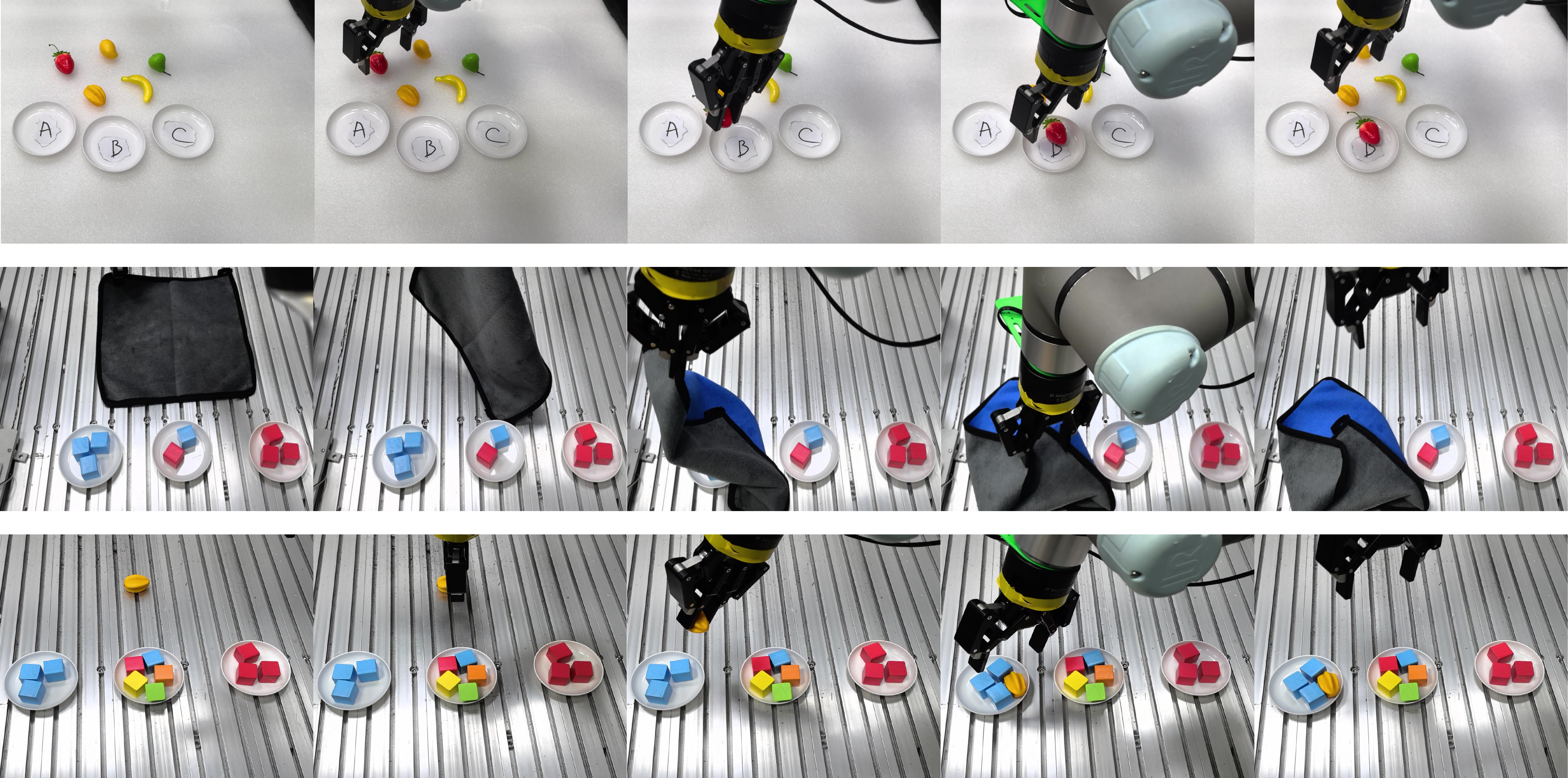}}
\caption{
The figure demonstrates the model’s performance on three ``visual reasoning'' tasks. 
For example, in the top sub-figure, we instruct the model to place a fruit into the plate labeled with the letter ``B''; 
in the middle sub-figure, the instruction is to cover the plate containing the most blue blocks with a towel; 
and in the bottom sub-figure, we require the model to place the fruit onto the plate with the highest number of blue blocks.
}\label{fig:appendix_task9}
\end{center}
\vspace{-20pt}
\end{figure*}

\begin{figure*}[t]
\begin{center}
\centerline{\includegraphics[width=1.6\columnwidth]{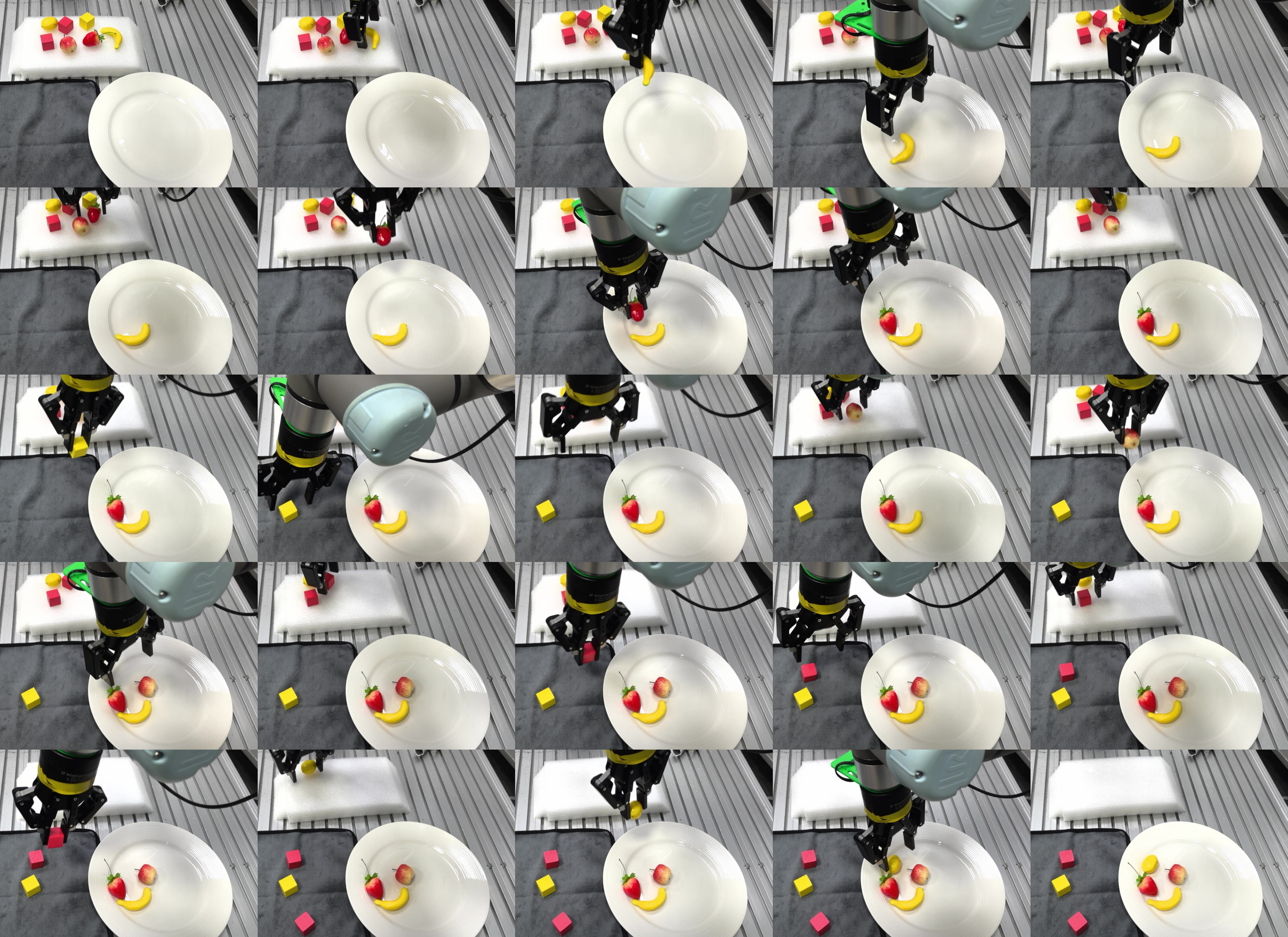}}
\caption{
In this long-horizon task, we require the model to classify objects by their type—for example, placing blocks on the towel and fruits in the plate.
}\label{fig:appendix_task10}
\end{center}
\vspace{-20pt}
\end{figure*}

\begin{figure*}[t]
\begin{center}
\centerline{\includegraphics[width=1.6\columnwidth]{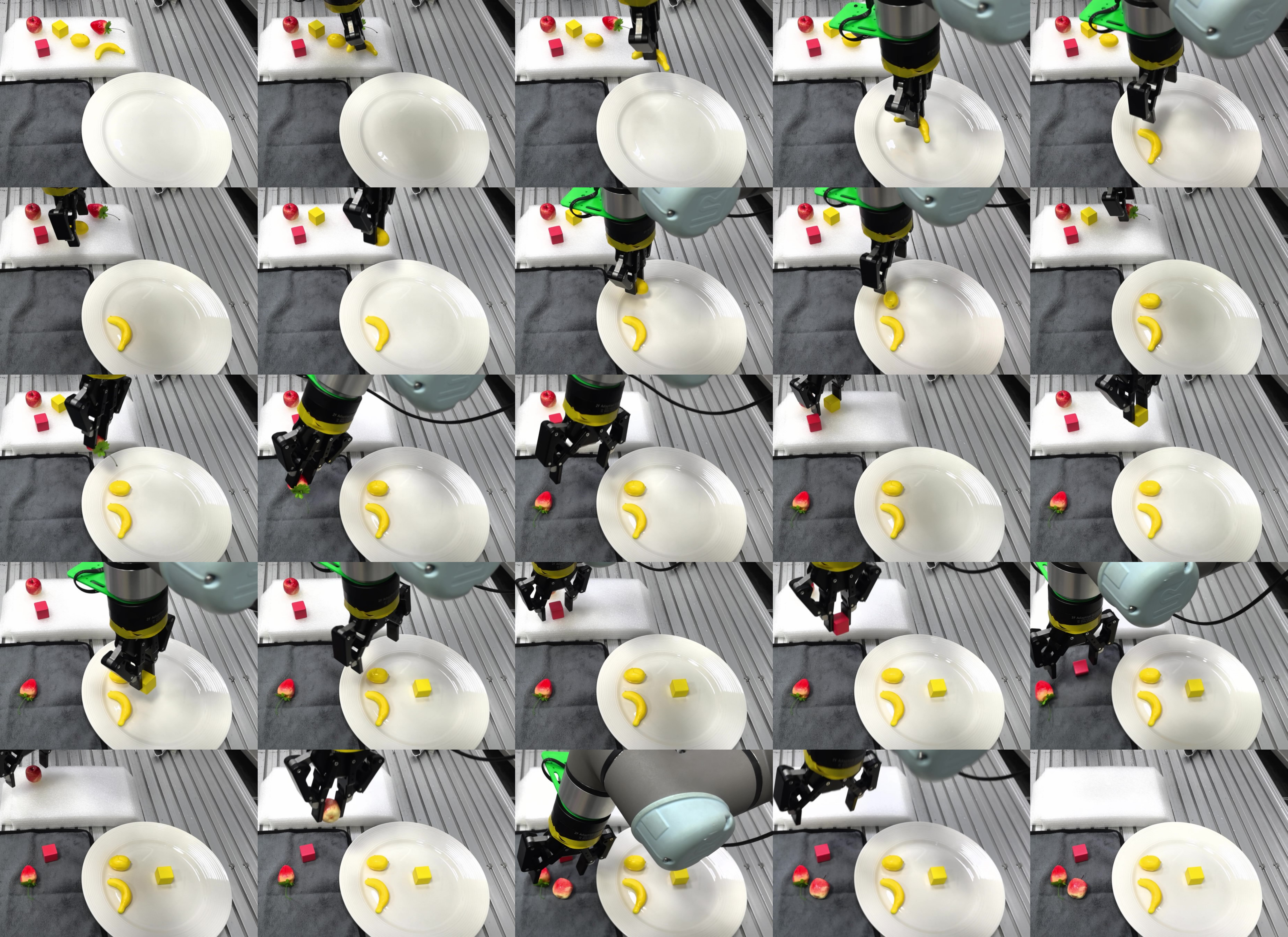}}
\caption{
In this long-horizon task, we require the model to classify objects by their color—for example, placing red objects on the towel and yellow objects in the plate.
}\label{fig:appendix_task11}
\end{center}
\vspace{-20pt}
\end{figure*}

In addition to quantitative results and analysis, we provide further visualizations in Figures~\ref{fig:appendix_task2}-\ref{fig:appendix_task11} to demonstrate the reliability of deploying the VITA model in real-world scenarios. We analyze our real-world results as follows:

(1) In complex scenarios, the Internal CoT enables the model to capture three-dimensional spatial features, allowing it to generate precise action trajectories in 3D space and successfully complete challenging tasks such as ``stacking multiple plates together.''

(2) The visual context empowers the model to plan and search for potential solutions by leveraging longer-term historical spatial information—for example, ``the model records the initial spatial relationships between objects and restores them to their original positions after task completion.'' This capability is critical for home environments, such as a robot returning jam and utensils to their original locations after assembling a sandwich.

\begin{figure*}[t]
\begin{center}
\centerline{\includegraphics[width=2.0\columnwidth]{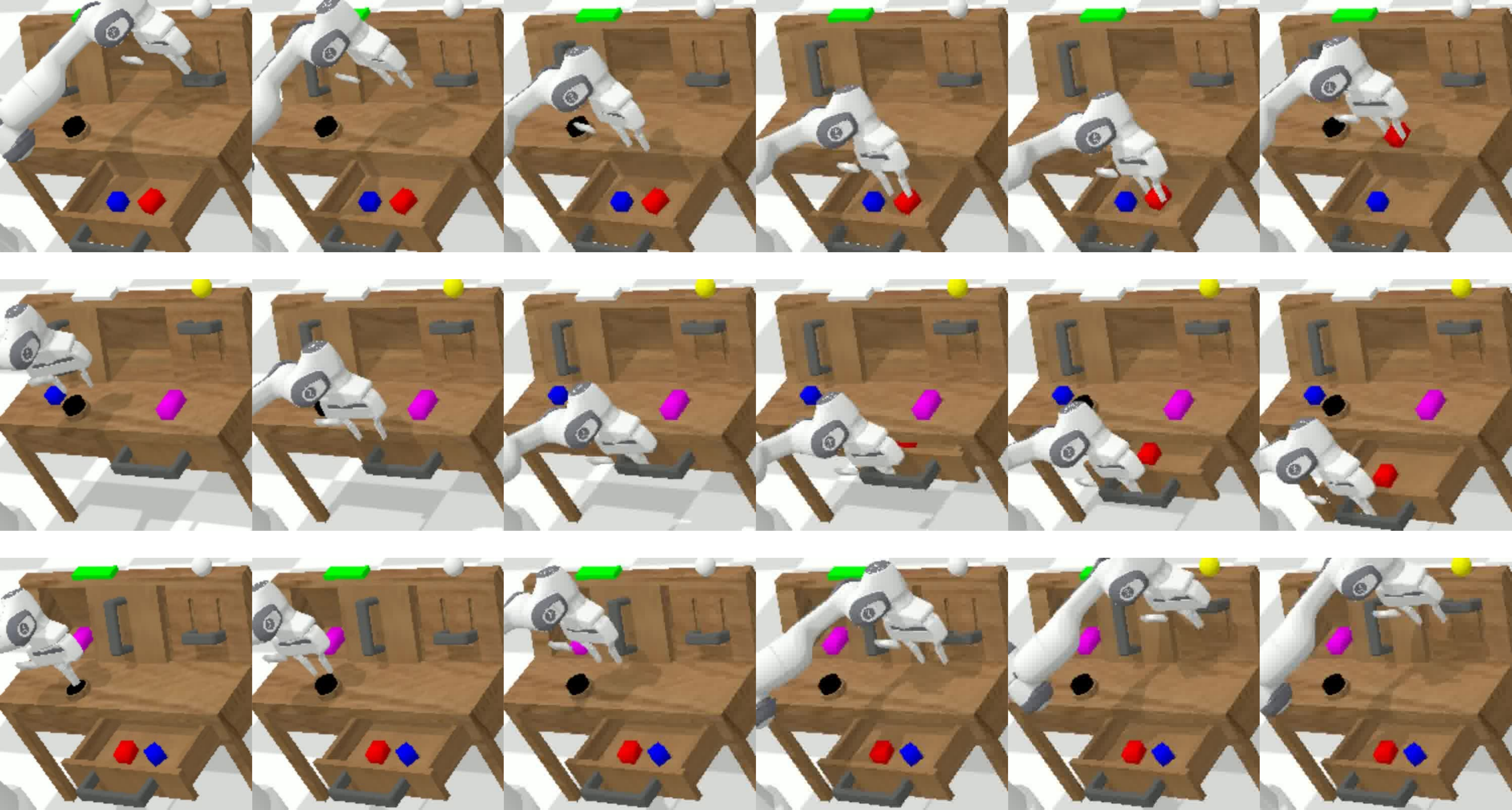}}
\caption{
Visualization of action trajectory generation by VITA in the CALVIN simulation environment.
}\label{fig:appendix_simu_calvin}
\end{center}
\vspace{-20pt}
\end{figure*}

\begin{figure*}[t]
\begin{center}
\centerline{\includegraphics[width=2.0\columnwidth]{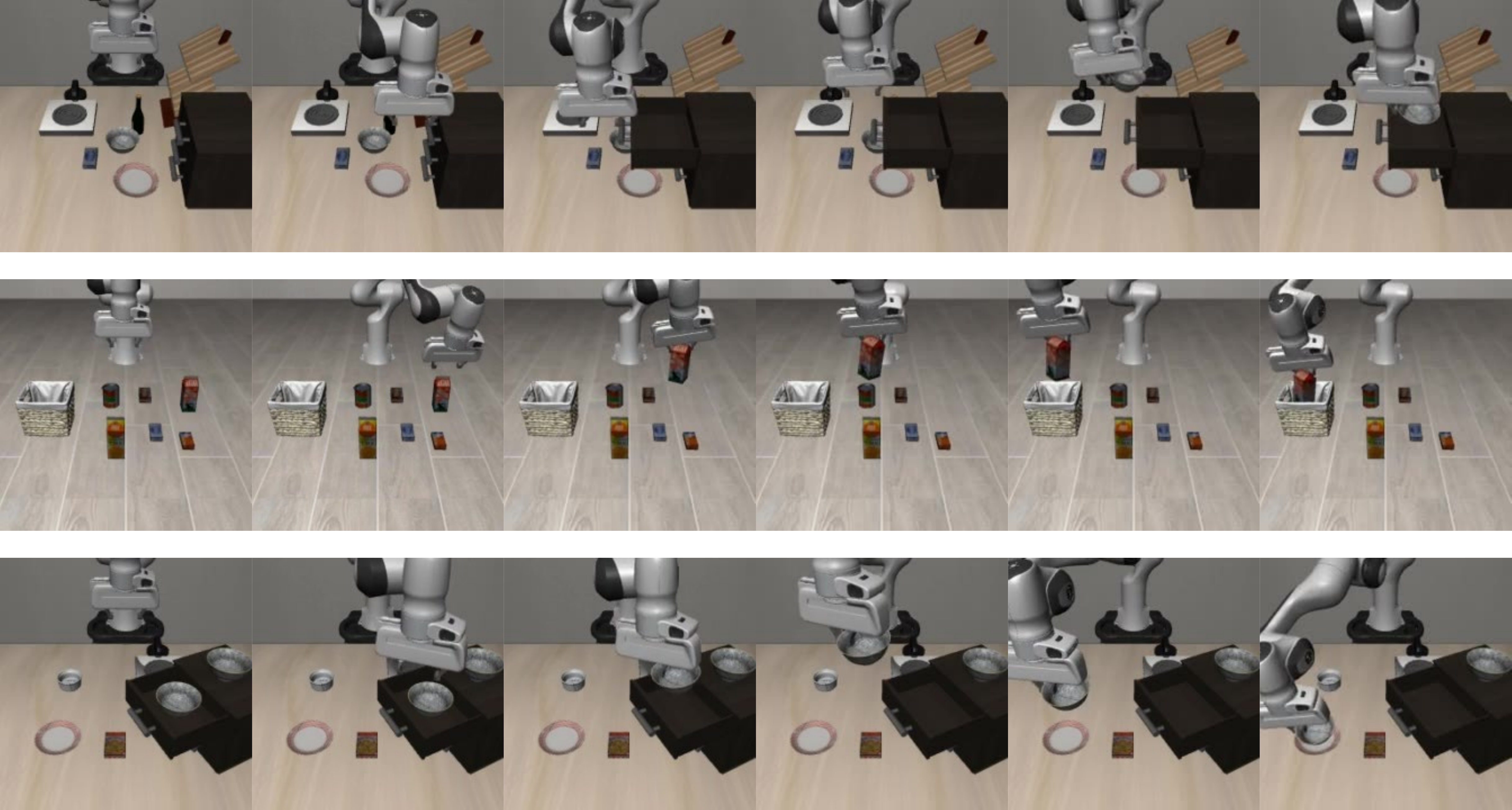}}
\caption{
Visualization of action trajectory generation by VITA in the LIBERO simulation environment.
}\label{fig:appendix_simu_libero}
\end{center}
\vspace{-20pt}
\end{figure*}

\begin{figure*}[t]
\begin{center}
\centerline{\includegraphics[width=2.0\columnwidth]{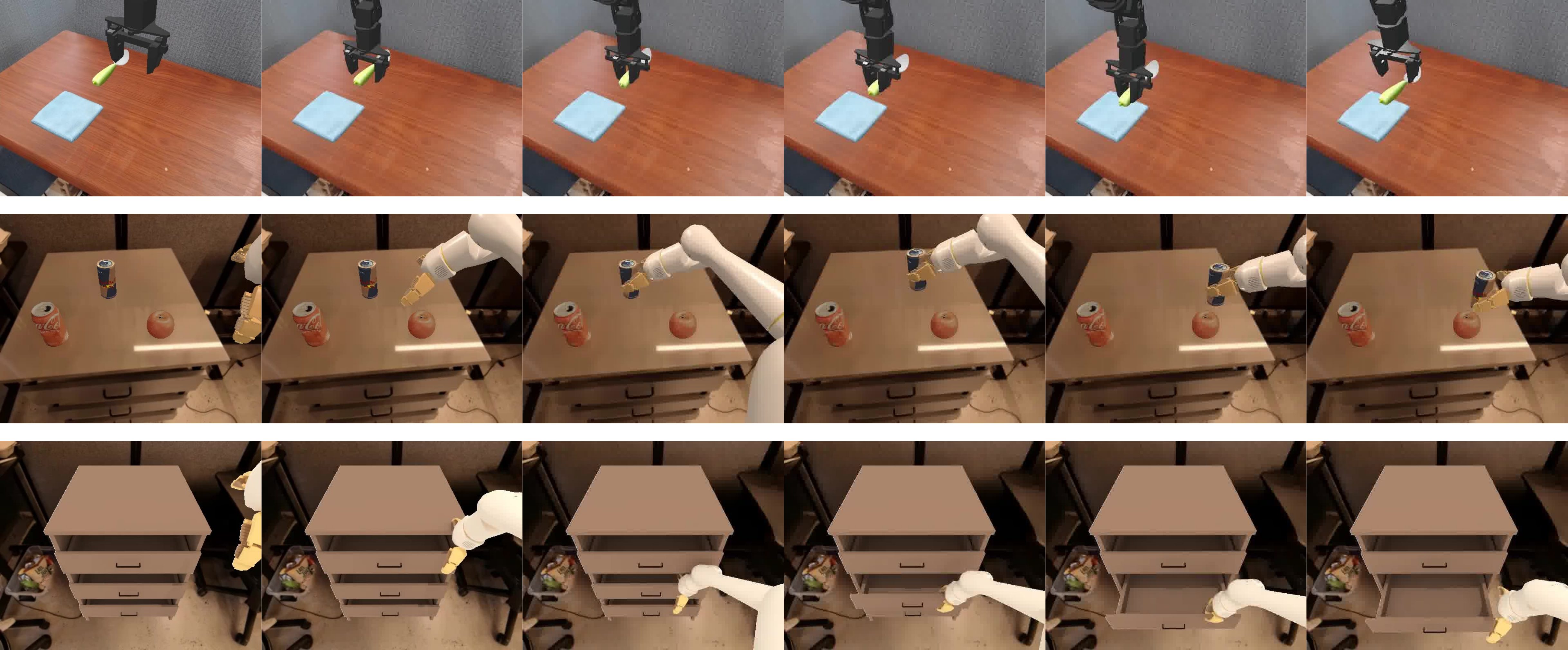}}
\caption{
Visualization of action trajectory generation by VITA in the SimplerEnv Google Robot and WidowX simulation environment.
}\label{fig:appendix_simu_simplerenv}
\end{center}
\vspace{-20pt}
\end{figure*}

Additionally, we present visualizations of VITA's generated action trajectories under simulation environments CALVIN, LIBERO, SimplerEnv Google Robot and WidowX in Figures~\ref{fig:appendix_simu_calvin}-\ref{fig:appendix_simu_simplerenv}.

\subsection{Visualization of the Ablation Model}
\begin{figure*}[t]
\begin{center}
\centerline{\includegraphics[width=2.0\columnwidth]{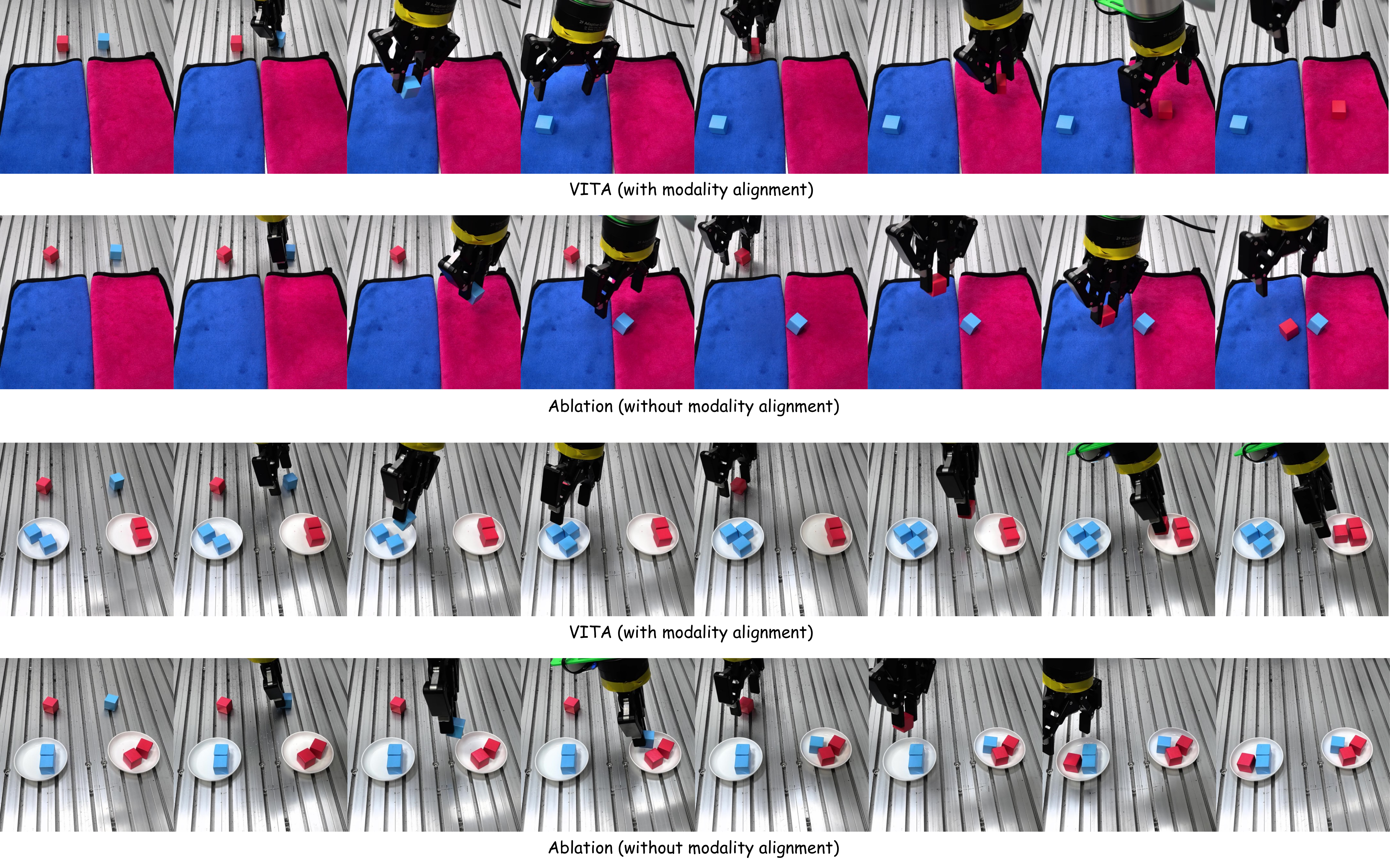}}
\caption{We visualize the performance of the model and its ablated variants in real-world ``Color Matching'' tasks.
}\label{fig:appendix_task12}
\end{center}
\vspace{-20pt}
\end{figure*}

Figure~\ref{fig:appendix_task12} presents visual comparisons from our ablation studies. Specifically, we design an ablated variant “w/o modality alignment”: this variant removes the shared codebook, skips the warmup stage, and proceeds directly to co-training and fine-tuning.

\subsection{Real-Robot Latency}\label{sec:latency}
In the real-world experiments, the UR-5e robotic platform receives two image observations, each with a resolution of 256×256 pixels, and outputs discrete action chunks over a horizon of 15 time steps, where each chunk contains 12 consecutive action steps. 
Each inference step takes approximately 2 seconds on an NVIDIA RTX 4090 GPU (24GB), and considering communication and data I/O overhead, the overall system latency is around 3 seconds.
The system operates at an action-chunk frequency of 5 Hz, while maintaining an average action-level inference throughput close to 60 Hz (i.e., number of action steps generated per second). 
This demonstrates that despite leveraging large-scale pretraining, VITA remains efficient for deployment on a single, commercially available computing device, without requiring extensive computational resources.

\section{Discussion}
\subsection{Limitations}
Although we have unified visual perception and action generation, our current framework does not fully integrate the visual chain-of-thought (V-CoT) with the textual chain-of-thought (T-CoT). 
Consequently, in complex, multi-step tasks that require deep textual reasoning and instruction grounding, such as ``making a sandwich''. 
Our method may struggle to fully capture the user's high-level intent. Nevertheless, VITA achieves remarkable advances in spatial scene modeling, motion coherence, and inference efficiency, demonstrating strong performance across a wide range of robotic manipulation benchmarks.

\subsection{Future Works}
A promising direction for future work lies in redesigning the discrete tokenization of action sequences to more faithfully preserve fine-grained temporal dynamics. 
Currently, VITA encodes action trajectories by first applying the Discrete Cosine Transform (DCT) to compress them into the frequency domain, followed by quantization via a shared codebook. 
While this strategy effectively reduces redundancy, the fixed basis functions of DCT are ill-suited for capturing non-stationary or aperiodic motion patterns commonly found in real-world tasks, such as abrupt stops, multi-phase maneuvers, or reactive corrections. 
More critically, this approach treats the entire action window as a monolithic unit, sacrificing precise time-step alignment between visual context and motor output. 
This limitation can be particularly detrimental in real-robot settings that demand millisecond-level timing precision for safe and robust execution.


\end{document}